\documentclass[lettersize,journal]{IEEEtran}
\usepackage{amsmath,amsfonts}
\usepackage{algorithmic}
\usepackage{algorithm}
\usepackage{array}
\usepackage[caption=false,font=normalsize,labelfont=sf,textfont=sf]{subfig}
\usepackage{textcomp}
\usepackage{stfloats}
\usepackage{url}
\usepackage{verbatim}
\usepackage{graphicx}
\usepackage{cite}
\usepackage{booktabs}
\usepackage{enumitem}

\usepackage[caption=false,font=normalsize,labelfont=sf,textfont=sf]{subfig}
\usepackage{ulem}
\usepackage{color}
\usepackage{pifont}
\usepackage{bbding}
\usepackage{multirow}
\usepackage{tabularray}
\usepackage{booktabs}

\usepackage{pdflscape}
\usepackage{rotating}
\usepackage{xspace}
\newcommand{\RADN}{RADN\xspace}
\newcommand{\HATSNet}{HATSNet\xspace}
\newcommand{\FBHR}{FBHR\xspace}
\newcommand{\TotalReLighting}{TR\xspace}
\newcommand{\SwitchLight}{SL\xspace}
\newcommand{\AFHIR}{AFHIR\xspace}
\newcommand{\Ours}{HumanMaterial\xspace}
\newcommand{\OursDataset}{OpenHumanBRDF\xspace}

\newcommand{\GeometryPriorModel}{Geometry Prior Model\xspace}
\newcommand{\GeometryPriorModelShort}{GPM\xspace}
\newcommand{\RSSPriorModel}{RSS Prior Model\xspace}
\newcommand{\RSSPriorModelShort}{RPM\xspace}
\newcommand{\AlbedoPriorModel}{Albedo Prior Model\xspace}
\newcommand{\AlbedoPriorModelShort}{APM\xspace}
\newcommand{\FinetuneModel}{Finetuning Model\xspace}
\newcommand{\FinetuneModelShort}{FTM\xspace}

\newcommand{\ControlledPBRRendering}{Controlled PBR Rendering\xspace}
\newcommand{\ControlledPBRRenderingShort}{CPR\xspace}

\newcommand{\etal}[0]{\textit{~et al.~}}

\begin{document}
	
\title{\Ours: Human Material Estimation from a Single Image via Progressive Training}

\author{Yu Jiang, Jiahao Xia, Jiongming Qin, Yusen Wang, Tuo Cao, and Chunxia Xiao${^{*}}$~\IEEEmembership{Senior Member,~IEEE,}
\thanks{
This work is partially supported by the National Natural Science Foundation
of China (No. 62372336).}
\thanks{Yu Jiang,  Jiongming Qin, Yusen Wang, Tuo Cao, and Chunxia Xiao are with the
	School of Computer Science, Wuhan University, Wuhan, China (e-mail:
	jiangyu1181@whu.edu.cn, 
	jiongming@whu.edu.cn, 
	wangyusen@whu.edu.cn,
	maplect@whu.edu.cn,
	cxxiao@whu.edu.cn).}
\thanks{Jiahao Xia is with the Faculty of Engineering and IT, University of Technology Sydney Ultimo, NSW, 2007, Australia (e-mail: jiahao.xia-1@uts.edu.au)
}
\thanks{${^{*}}$Chunxia Xiao is the corresponding author.}
}


\maketitle

\begin{abstract}
\label{abstract}
Full-body Human inverse rendering based on physically-based rendering aims to acquire high-quality materials, which helps achieve photo-realistic rendering under arbitrary illuminations. This task requires estimating multiple material maps and usually relies on the constraint of rendering result. The absence of constraints on the material maps makes inverse rendering an ill-posed task.
Previous works alleviated this problem by building material dataset for training, but their simplified material data and rendering equation lead to rendering results with limited realism, especially that of skin. 
~To further alleviate this problem, we construct a higher-quality dataset~(\OursDataset) based on scanned real data and statistical material data. In addition to the normal, diffuse albedo, roughness, specular albedo, we produce displacement and subsurface scattering to enhance the realism of rendering results, especially for the skin. 
With the increase in prediction tasks for more materials, using an end-to-end model as in the previous work struggles to balance the importance among various material maps, and leads to model underfitting. Therefore, we design a model~(\Ours) with progressive training strategy to make full use of the supervision information of the material maps and improve the performance of material estimation. 
\Ours first obtain the initial material results via three prior models, and then refine the results by a finetuning model. Prior models estimate different material maps, and each map has different significance for rendering results.
Thus, we design a \ControlledPBRRendering (\ControlledPBRRenderingShort) loss, which enhances the importance of the materials to be optimized during the training of prior models.
~Extensive experiments on \OursDataset dataset and real data demonstrate that our method achieves state-of-the-art performance. 
\end{abstract}


\section{Introduction}
\label{introduction} 
Physically Based Rendering (PBR) is a ray-tracer rendering technique in computer graphics that leverages physically accurate material definitions (i.e., PBR materials) to achieve photo-realistic rendering under arbitrary illuminations. This technique is widely applied in video games, virtual reality, 3D printing, and animated movies.

Inverse rendering is a technique widely used to obtain human PBR materials. Its core lies in inferring the material maps of an object from the rendering results. Different combinations of PBR materials may produce similar rendering results, and some of these combinations may even be obviously wrong. Moreover, the lack of key information in the input image data (such as only a single image) and the simplification of the PBR rendering model will exacerbate the ambiguity in material estimation. The above issues make human inverse rendering an ill-posed task.
 
An effective way to solve this problem is to employ supervision information to constrain the estimated materials. Previous works~\cite{TotalRelighting_2021_Sig_single_image_human_phong_neural_render,Lagunas_2021_EGSR_single_image_human_neural_render,AFHIR_Daichi_2024_arxiv,ICLight,SwitchLight_2024_CVPR_pbr_and_neural_render} have attempted to construct human material datasets to alleviate this task. Among them, \TotalReLighting~\cite{TotalRelighting_2021_Sig_single_image_human_phong_neural_render} constructed a high-quality dataset based on real capture data, but the simplified types of material maps and the simplified rendering model limit the applications. \FBHR~\cite{Lagunas_2021_EGSR_single_image_human_neural_render}, on the other hand, used a rendering engine to construct synthetic data. More types of PBR material maps have expanded the applications, but the realism of rendering results is limited.
The above datasets are hard to meet the requirements of higher-quality PBR rendering, and are struggling to represent realistic skins.
\begin{figure}[!t]
	\centering 
	\begin{minipage}[b]{\linewidth} 
		\centering 
		\newcommand{\myvspace}{1.0 pt} 
		\newcommand{\widthOfFullPage}{1} 
		\newcommand{\widthOfMiniPage}{0.95}
		\newcommand{\format}{png}
		\subfloat{
			\begin{minipage}[b]{\widthOfFullPage\linewidth} 
				\centering
				\includegraphics[width=\widthOfMiniPage\linewidth]{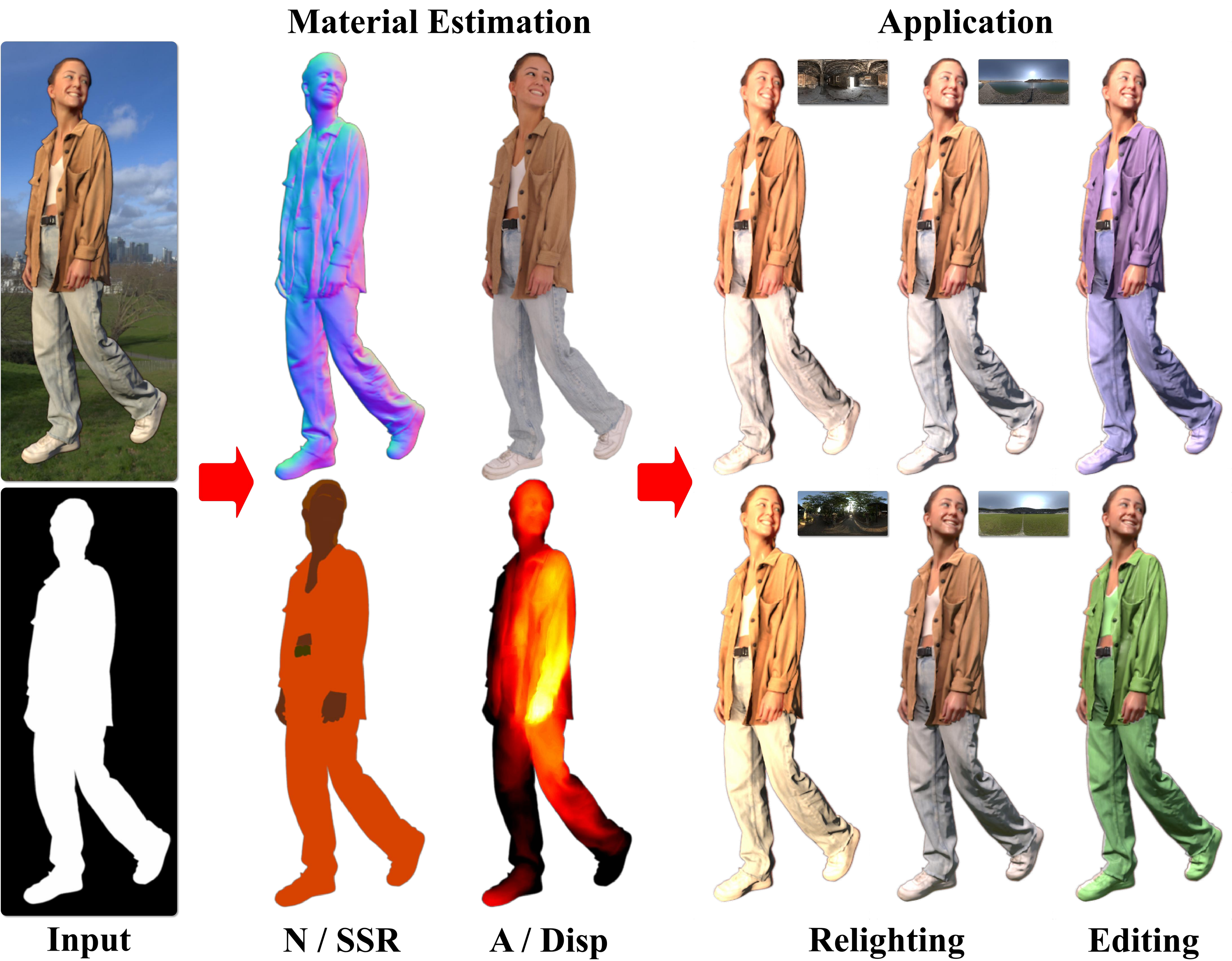}
			\end{minipage}
		}
	\end{minipage}
\caption{
 \Ours. Our method aims to estimate PBR materials from a single human image and foreground mask. It enables the estimation of PBR materials, including ``N" (normal),  ``A" (diffuse albedo), ``SSR" (subsurface scattering, specular albedo, and roughness) and ``Disp" (displacement). Leveraging these estimated results, our method facilitates realistic applications such as relighting and material editing under a novel environment map. 
	}
\label{img-teaser}
\end{figure} 

To further alleviate this ill-posed task, we produce a higher-quality human PBR material dataset. The statistical data of various materials are demonstrated in \cite{RTR_2018_Akenine}, among which those related to the human body are hair, skin, fabric, and leather. Based on this, we manually design the material parameters and produce \OursDataset in Blender~\cite{Blender} based on the statistical data. The dataset includes 147 human models and covers six ethnicities.
In addition to the widely used material maps (normal, diffuse albedo, roughness, and specular albedo), we also produce displacement and subsurface scattering data.
The displacement map shows the surface and helps to make more detailed rendering results when there is no complete geometric information. The subsurface scattering map can enhance skin realism by using advanced PBR shaders. This makes the visual effect of the presented skin more realistic and closer to the real world. 

To achieve realistic rendering, we build more material data and employ an advanced PBR shader, which results in an increase in the number of material maps that the model needs to estimate. As shown in Fig.~\ref{img-teaser}, this model need to estimate six material maps, so as to make applications such as relighting and material editing. Previous works that use an end-to-end model to estimate all material maps simultaneously will exacerbate the difficulty of model training and limit the performance of material estimation. Therefore, to make full use of the supervision information of the material maps, we design the \Ours with a progressive training strategy. 

\Ours first estimates six material maps using three prior models, and then employs a finetuning model to jointly optimize all materials. The prior model trained independently can improve the performance of material estimation. The finetuning operation enhances the connection between material maps through PBR with various illuminations constraints, and helps to obtain more convincing material maps. 

Based on the correlations between material maps, we carefully design the target map for each prior model estimation. Specifically, the \GeometryPriorModel estimates normal and displacement, \AlbedoPriorModel estimates diffuse albedo, and \RSSPriorModel estimates roughness, specular albedo, subsurface scattering. 
Furthermore, the significance of each material map to the rendering outcome differs. To increase its significance, we design a \ControlledPBRRendering loss function. 
The key to this rendering loss lies in carefully setting the values of non-optimized materials within a reasonable range not just using the GT data. By doing so, it effectively mitigates the influence that non-optimized parameters exert on the rendering results across various illuminations. This approach also makes the training process more stable and the estimated results more accurate. This makes the estimated material maps easier to be optimized during the training process.

To summarize, we make the following main technical contributions.
\begin{itemize}
	\item Produce a higher-quality Human PBR material dataset (\OursDataset), which has well-designed parameter setting and realistic skin representation. 
	\item Propose a material estimation method (\Ours) from a single human image, which constructs three prior models to obtain the initial material, and estimates higher-quality PBR materials after finetuning.
	\item Design a \ControlledPBRRendering loss function, which can enhance the significance of material to be optimized during training, thereby improving the performance of prior models.
\end{itemize}

Experiments on our dataset and real data demonstrate that our method can estimate high-quality materials, achieve realistic relighting, and attain state-of-the-art performance.

\section{Related work}
\label{related_work}
Inverse rendering is a highly regarded research topic in computer graphics. In previous works, some methods are based on simplified PBR models~\cite{TotalRelighting_2021_Sig_single_image_human_phong_neural_render,Photorealistic_2022_CVPR_single_image_human_neural_render,Unsupervised_Learning_Liu_2020_CVPR} for appearances decomposition. Others employ the neural-based render to achieve relighting~\cite{Lagunas_2021_EGSR_single_image_human_neural_render,Photorealistic_2022_CVPR_single_image_human_neural_render,TotalRelighting_2021_Sig_single_image_human_phong_neural_render,Single_image_portrait_relighting_2019_SIGGRAPH}.
For more robust relighting and material editing under complex lighting conditions, some works focus on estimating PBR materials~\cite{Li_2018_TOG_single_image_natural_svbrdfs,NeRO_2023_Sig_Multi_images_natural_svbrdf, SwitchLight_2024_CVPR_pbr_and_neural_render}. 
~These methods can be categorized based on the input data, including single-image-based~\cite{RADN_2018_TOG_single_image_natural_svbrdf,HATSNet_2021_TOG_single_image_natural_svbrdf}, multi-images-based~\cite{PhySG_2021_CVPR_multi_images_natural_brdf,NeILF_2022_ECCV_Multi_images_natural_svbrdf,MaXiaoHe_TVCG_2024,BRDF_Acquisition_TVCG_2024,10929734}, and video-based approaches~\cite{Relightable_2023_ICCV_video_human_svbrdf}.

\subsection{Appearance Decomposition}
\label{appearance_decomposition} 
\noindent\textbf{Single-image Human Appearance Decomposition.} 
For single human image relighting, an effective way is using the empirical model to decompose appearance. For example, Lagunas\etal\cite{Lagunas_2021_EGSR_single_image_human_neural_render} presents a data-driven method to decompose a single full-body image to diffuse, specular, and a light-dependent residual term.
Pandey\etal\cite{TotalRelighting_2021_Sig_single_image_human_phong_neural_render} decomposes appearance into diffuse, specular components and trains a neural render to achieve relighting.
PhorHum~\cite{Photorealistic_2022_CVPR_single_image_human_neural_render} is an end-to-end trainable deep neural network for photorealistic 3D human reconstruction via a monocular RGB image. 

\textbf{Multi-images Appearance Decomposition.} Unsupervised appearance representation can be achieved based on multi-images and neural radiation fields. 
Based on a data-driven BRDF prior and NeRF, NeRFactor~\cite{Nerfactor_2021_SigAsia_multi_images_natural_brdf} recovers object's shape and spatially-varying reflectance from the posed multi-view images. PhySG~\cite{PhySG_2021_CVPR_multi_images_natural_brdf} represents specular BRDFs and environmental illumination using mixtures of spherical gaussians, and can reconstruct natural geometry, materials, and illumination from multi-images. 
~The above methods can obtain impressing relighting results. As they are hard to obtain PBR material, which limits the realism of rendering under complex lighting and makes it difficult to achieve class-level material editing.

\subsection{PBR Material Estimation and Reconstruction}
\noindent\textbf{Single-image Material Estimation.} 
Single-image material estimation is more challenging than appearance decomposition due to the ill-posed PBR. Some significant works~\cite{Li_2017_TOG_single_image_plane_natural_svbrdf,LATNet_2022_TOG_single_image_natural_svbrdf} have been published since the near-planar PBR material dataset~\cite{RADN_2018_TOG_single_image_natural_svbrdf} have been produced. 
~RADN~\cite{RADN_2018_TOG_single_image_natural_svbrdf} produces a large synthetic near-planar surface material dataset and used an unet to recover materials from a single near-planar capture. Guo \etal\cite{HATSNet_2021_TOG_single_image_natural_svbrdf} designed a highlight-aware convolutional module alleviating the artifact problem in the estimated materials caused by overexposed regions.
~Based on the pre-trained network, ~Guo \etal\cite{DIR_2019_TOG_single_image_natural_svbrdf} and Guo \etal~\cite{MaterialGAN_2020_TOG_single_image_natural_svbrdf} reduced the error between the reconstructed result and the appearance (input) by optimizing the network in the latent space. 


\textbf{Multi-images Material Reconstruction.} 
Lots of works~\cite{InvRender_2022_CVPR_Multi_images_natural_svbrdf,NeILF_2022_ECCV_Multi_images_natural_svbrdf,Relightify_2023_ICCV_DiffusinModel_SVBRDF_facial,TensoIR_2023_CVPR_Multi_images_natural_svbrdf,TenseRF_2022_ECCV,NeRD_2021_ICCV_Multi_images_natural_svbrdf} have tried reconstructing PBR material  from multiple images.
~For example, Nvdiffrec~\cite{Nvdiffrec_2022_CVPR_natural_multi_images_svbrdf} leverages coordinate-based networks to represent volumetric texturing compactly, uses differentiable marching tetrahedrons to enable gradient-based optimization on the surface.
Specular reflections are view-dependent and break the multi-view consistency, which makes it challenging to reconstruct geometry and materials of reflective objects~\cite{Factored_NeuS_2023_arxiv_Multi_images_natural_svbrdf}. To address this challenge, NeRO~\cite{NeRO_2023_Sig_Multi_images_natural_svbrdf} is a two-step approach, which first reconstruct the geometry and then recover environment lights and the material. 

\textbf{Multi-images Human Material Reconstruction.} 
Recently, there have been some works~\cite{AvatarMe++_2022_TPAMI_Lattas,AvatarMe_2020_CVPR_Lattas,Shu_Hadap_Shechtman_Sunkavalli_Paris_Samaras_2017} attempting to estimate the human-related PBR materials. Relit-NeuLF~\cite{Relit_NeuLF_2023_ACMMM_Multi_images_facial_svbrdf} leverages a two-plane light field representation to parameterize each ray in a 4D coordinate system for efficient learning and inference, then recovers facial materials from multi-view images in a self-supervised manner.
After capturing the facial images with cross-polarized and parallel-polarized light with an inexpensive polarization foil, Azinovi\etal\cite{HighRes_2023_CVPR_images_facial_svbrdf} reconstructs explicit surface mesh of the face using structure from motion and then exploit the camera and light co-location within a differentiable render to optimize the facial materials using an analysis-by-synthesis approach.
~While differentiable inverse rendering-based methods have succeeded in static objects, some impressive works~\cite{Relighting4D_2022_ECCV_video_human_svbrdf,Relightable_2023_ICCV_video_human_svbrdf,zhen_2023_Arxiv_video_human_svbrdf} attempt to perform material estimation on dynamic human.

Above methods have achieved impressive results in material reconstruction, while for the human image with diverse material categories, material reconstruction without category-guided distinction may lack accuracy. Furthermore, it can result in insufficient differences in the realism of rendering results, particularly in fabric, leather, and skin.
~Benefiting from our PBR material dataset with classification, our method can estimate more accurate PBR materials and achieve realistic relighting and material editing.
\begin{table}[!ht]
\centering
\caption{
Comparison with previous dataset (\FBHR~\cite{Lagunas_2021_EGSR_single_image_human_neural_render}, \AFHIR~\cite{AFHIR_Daichi_2024_arxiv}, and \TotalReLighting~\cite{TotalRelighting_2021_Sig_single_image_human_phong_neural_render}). ``Open" means the dataset is open-sourced. ``Type" means the data type, and the ``Scan" means scanned form real data.
~``Render" means the supported renderer types of the data. ``Geo" means the geometric data in dataset. ``Skin" means whether the dataset enhance skin realism.
}
\label{table-dataset_compare}
\begin{tblr}{
		cells = {c},
		hline{1-2,6} = {-}{},
	}
	& Open & Type      & Render & Geo & Skin \\
	\FBHR  & /       & Synthetic & Neural & /        & /    \\
	\AFHIR & /       & Synthetic & PBR+Neural    & Depth        & /    \\
	\TotalReLighting & /       & Scan~   & Neural & /        & /    \\
	Ours  & \checkmark & Scan~   & PBR    & Surface    & \checkmark  
\end{tblr}
\end{table}

\section{\OursDataset Dataset}
\label{dataset_building}
Previous works have built some material datasets, as shown in Table~\ref{table-dataset_compare}, both \FBHR~\cite{Lagunas_2021_EGSR_single_image_human_neural_render} and \AFHIR~\cite{AFHIR_Daichi_2024_arxiv} build a dataset for estimating human PBR materials from a single image. However, these data are synthetic and have a fixed gap from the real scene.
~\TotalReLighting~\cite{TotalRelighting_2021_Sig_single_image_human_phong_neural_render} builds a real dataset for portrait relighting, the material data include normal and diffuse albedo, and does not support PBR render. It should be noted that none of these datasets are open-source, and we will release our dataset once this paper is accepted. 

Producing high-quality PBR material datasets in industry is important and commercially valuable. The ill-posed problem of PBR inverse rendering usually demands that experienced artists make tedious adjustments. Besides, the materials in human body are complex, and the appearances of different individuals vary greatly. All these factors induce the absence of high-quality open-source human PBR material datasets. 

\begin{table}[!t]
\centering
\caption{Value setting of the specular albedo (Specular), roughness (Roughness), and subsurface scattering (SSS).}
\begin{tabular}{l|c|c|c}
   \toprule
   Dielectric & Specular & Roughness & SSS\\
   \midrule
   Hair &   0.239&   0.500&   0.00\\
   Skin &   0.184&   0.400&   0.08\\
   Fabric & 0.263&   0.850&   0.00\\
   Leather &   0.224&   0.250&   0.00\\
   \bottomrule
\end{tabular}
\label{table-f0_roughness_value_range}
\end{table}
\begin{figure}[!h]
  \centering 
  \begin{minipage}[b]{\linewidth} 
  \newcommand{\myvspace}{1.0 pt} 
  \newcommand{\widthOfFullPage}{1} 
  \newcommand{\widthOfMiniPage}{1}
  \newcommand{\format}{png}
  \subfloat{
    \begin{minipage}[b]{\widthOfFullPage\linewidth} 
      \centering
    \includegraphics[width=\widthOfMiniPage\linewidth]{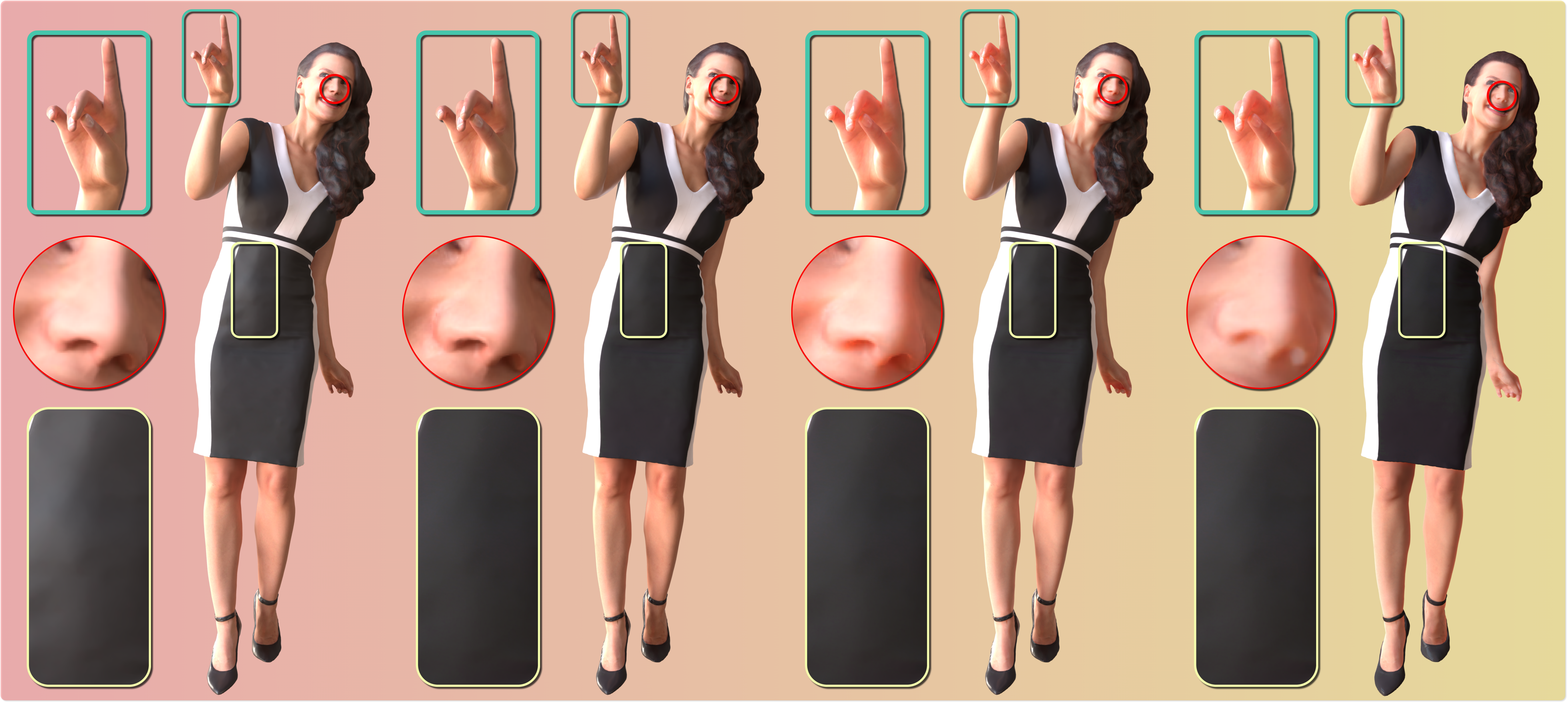}\vspace{\myvspace}   
    \end{minipage}
  }
  \begin{flushleft}
  	\small
  	\vspace{-0.5em}
	\hspace{1.8em}(a) Origin
	\hspace{2.6em}(b) Adjust
	\hspace{1.5em}(c) Adjust+SSS
	\hspace{1.5em}(d) Surface
  \end{flushleft}
  \end{minipage}
\caption{
Data realism enhancement. (a) represents the original data we collected. (b) shows the result with the materials~(roughness and specular albedo) after we adjusted. (c) shows the result after adjustment and using subsurface scattering. (d) shows the result that we use the displacement data to represent the surface based on the adjusted materials of (c).
Please zoom in for the details and we analyze it in the Sec. \ref{dataset_building}.
}
\label{img-data_enhance}
\end{figure}
Based on a large amount of experimental data, RTR~\cite{RTR_2018_Akenine} calculates the value ranges of specular albedo of typical materials in the real world, and the human-related materials are hair, skin, fabric, and leather. In most cases, they are also the main materials of the full body. 
To make human PBR material data production feasible and to simplify the process, we make a reasonable assumption that the full-body mainly contains four material categories, namely hair, skin, fabric, and leather.
~Referring to the statistical ranges of specular albedo for these materials in RTR\cite{RTR_2018_Akenine}, we set a value for each material. Additionally, based on the approximate size relationship of the roughness for different materials (from small to large: leather, skin, hair, fabric), we adjust the roughness values of each material in Blender to ensure realistic appearance under different illuminations. The final settings are presented in Table~\ref{table-f0_roughness_value_range}. Besides, we collect 1092 HDR environment maps (782 of real world from~\cite{HDRI}, and 310 synthetic we build), and they are used for rendering appearance in Blender.

After confirming the parameters settings of the materials, we collect 147 high-quality human models from RenderPeople~\cite{RenderPeople}, the vanilla data are scanned from the real world. To cover a wide range of human diversity, we establish three standards. (1) We select models with a 50\% balance between males and females. (2) All human models come from six ethnicities, including Asian, Black, Indian, Middle Eastern, White, and Hispanic/Latino, with each ethnicity accounting for 16.7\%. (3) The age range covered is the largest among all age groups, 19-50. 

Based on the vanilla data and reasonable setting, we build the \OursDataset dataset in Blender. (1)~We label each human model's hair, skin, fabric, and leather regions. (2)~According to the above settings, we manually adjust each class's roughness, specular albedo, and subsurface scattering. (3)~We bake material UV maps for each human, including normal, diffuse albedo, roughness, specular albedo, subsurface scattering and displacement maps. (4)~We fix the camera, rotate the human model, and render appearances under two environment maps. 
\begin{figure*}[!ht]
  \centering 
  \begin{minipage}[b]{\linewidth} 
  \centering 
  \newcommand{\myvspace}{1.0 pt} 
  \newcommand{\widthOfFullPage}{1} 
  \newcommand{\widthOfMiniPage}{1}
  \newcommand{\format}{png}
  \subfloat{
    \begin{minipage}[b]{\widthOfFullPage\linewidth} 
      \centering
    \includegraphics[width=\widthOfMiniPage\linewidth]{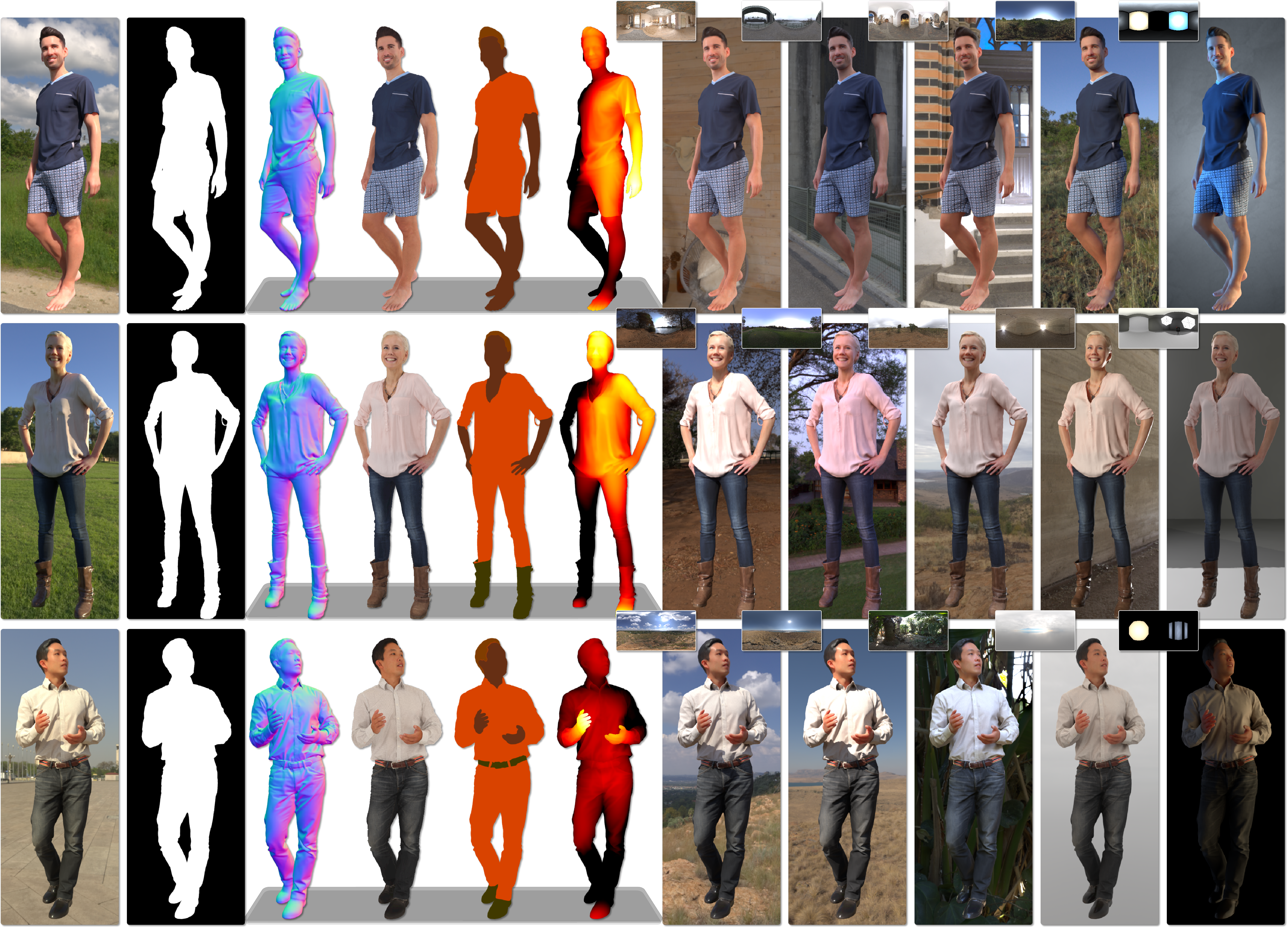}\vspace{\myvspace}
    \end{minipage}
  }
  \begin{flushleft}
  	\small
  	\vspace{-0.5em}
    \hspace{0.7em}Appearance
    \hspace{1.7em}Mask
    \hspace{8.5em}Materials
    \hspace{9.8em}Relightings under real(4) and synthetic(1) illuminations
  \end{flushleft}
  \end{minipage} 
\caption{Three samples of \OursDataset. ``Appearance" means the input image of model. In ``Materials", the normal, diffuse albedo, SSR (subsurface scattering, specular albedo, and roughness are concatenated in the channel axis), and displacement map are from left to right. We use gamma correction to brighten diffuse albedo for better vision. Please see more samples in the attached video.}
\label{img-human_sample}
\end{figure*}

With above processing, we can obtain the reasonable material maps, as well as render more realistic appearance compared to the result which only uses original collected data. 
~As shown in Fig. \ref{img-data_enhance}, the fabric (dress) in (a) has unreasonable highlights, making it look plastic and unrealistic. We can find (b) looks more realistic demonstrating that our PBR materials setting are accurate. In addition, by comparing (a), (b), and (c), we can see that the subsurface scattering in our materials enhances the realism of skin, especially on the nose and hand.

We produce 147 sets of data, 127 are used for training and 20 for testing. 
~For the training data, we render 100 image sets for each model. Therefore, the training set comprises a total of 12,700 data sets. When rendering the image set, the initial view is that the camera is facing the model directly. Then, we rotate the camera evenly around the model for one full circle and render 100 times to obtain multi-view rendering results.
~For the testing data, we render 10 image sets for each model under the initial view. Consequently, the test set has 200 data sets in total. 
~Three samples are shown in Fig.~\ref{img-human_sample}, the resolution of the image set is $512\times512$ or $4096\times2048$ (only environment map). Each set includes the foreground mask, PBR materials (normal, diffuse albedo, roughness, specular albedo, subsurface scattering, and displacement), and five relighting results under different environment maps. The displacement data presents the distance from the geometric surface to the world center plane.
~It should be noted that the boundaries of material categories may be ambiguous, such as hair and skin, which can cause minor classification errors.  That is why we do not simply consider the material estimation as a classification task. 

\begin{figure*}[!ht]
\centering
\includegraphics[width=\textwidth]{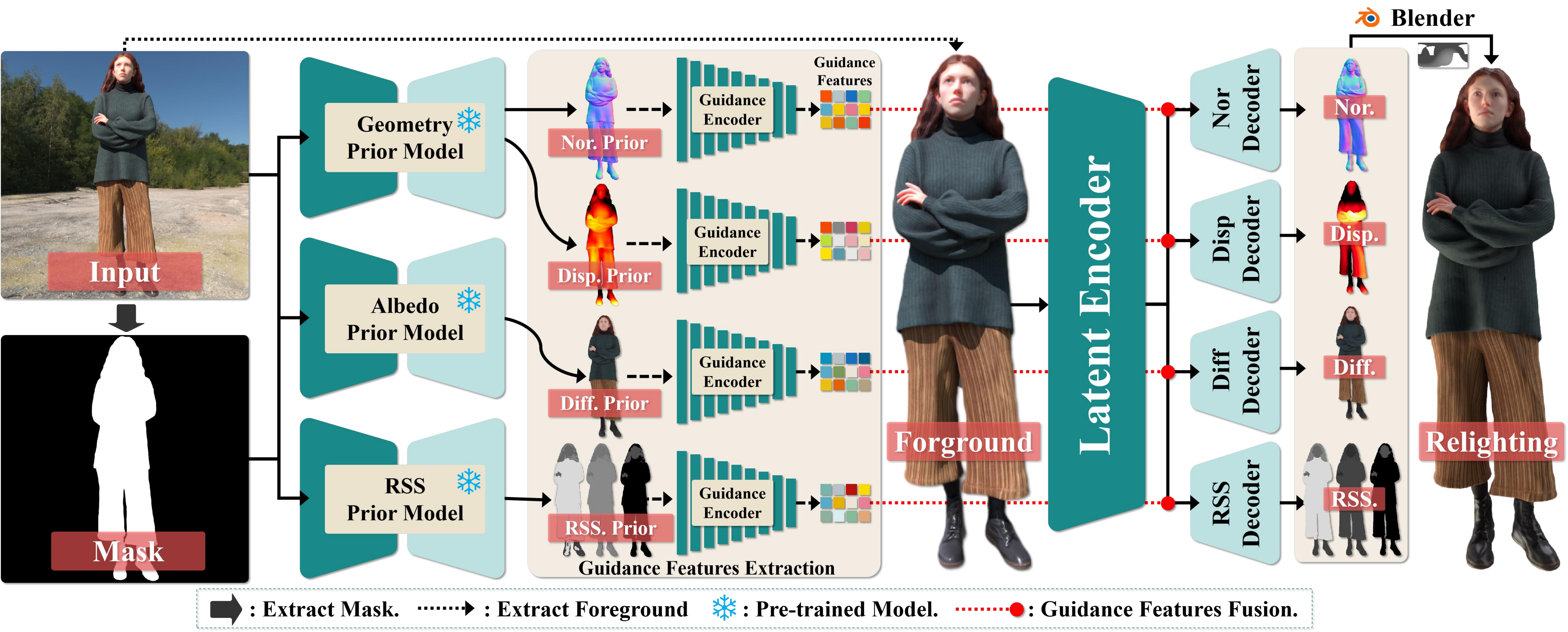}
\caption{
Method overview. Our method, \Ours, estimates the PBR materials from a single human image, and achieves photorealistic relighting result. 
~First, we use \cite{Rembg} to acquire ``Mask" and use alpha matting to extract ``Foreground". After building our our \OursDataset dataset (Sec. \ref{dataset_building}), we train three prior models (Sec. \ref{prior_models}) to estimate material priors. 
~The \GeometryPriorModel estimates normal (``Nor.") and displacement (``Disp.") priors. The \AlbedoPriorModel estimates the diffuse albedo (``Diff.") prior. The \RSSPriorModel estimates ``RSS" prior, which consists of roughness, specular albedo, and subsurface scattering.
~Based on the prior models, we build a Guidance Encoder to extract guidance features from the four priors. 
~Then, we design a \FinetuneModel (Sec. \ref{finetune_model}) which consists of a Latent Encoder and four (``Nor", ``Disp.", ``Diff.", and ``RSS" ) decoders to estimate the final PBR materials. The Material Model uses the Guidance Features to guide the decoding process, and after joint optimization, it estimates the final materials (``Nor.", ``Disp.", ``Diff.", and ``RSS."). The render-ready materials can be used in the ray-traceable Blender for relighting under arbitrary illuminations. 
}
\label{overview}
\end{figure*}
\section{Method}
\label{method}
Given a single human image, our task is to estimate PBR materials, which include normal, diffuse albedo, roughness, specular albedo, displacement and subsurface scattering. 
Estimating six material maps from a single image is a challenging task. To make full use of the material supervision information and make the training of the model fitting more easily, as shown  in Fig~\ref{overview}, we propose the \Ours with a progressively training strategy to handle this task. 
\Ours first uses three prior models to estimate six material priors. Each model estimates one, two, and three materials respectively. Subsequently, a joint optimization is used to refine the result.

\begin{figure}[!h]
	\centering 
	\begin{minipage}[b]{\linewidth} 
		\newcommand{\myvspace}{1.0 pt} 
		\newcommand{\widthOfFullPage}{1} 
		\newcommand{\widthOfMiniPage}{1}
		\newcommand{\format}{png}
		\subfloat{
			\begin{minipage}[b]{\widthOfFullPage\linewidth} 
				\centering
				\includegraphics[width=\widthOfMiniPage\linewidth]{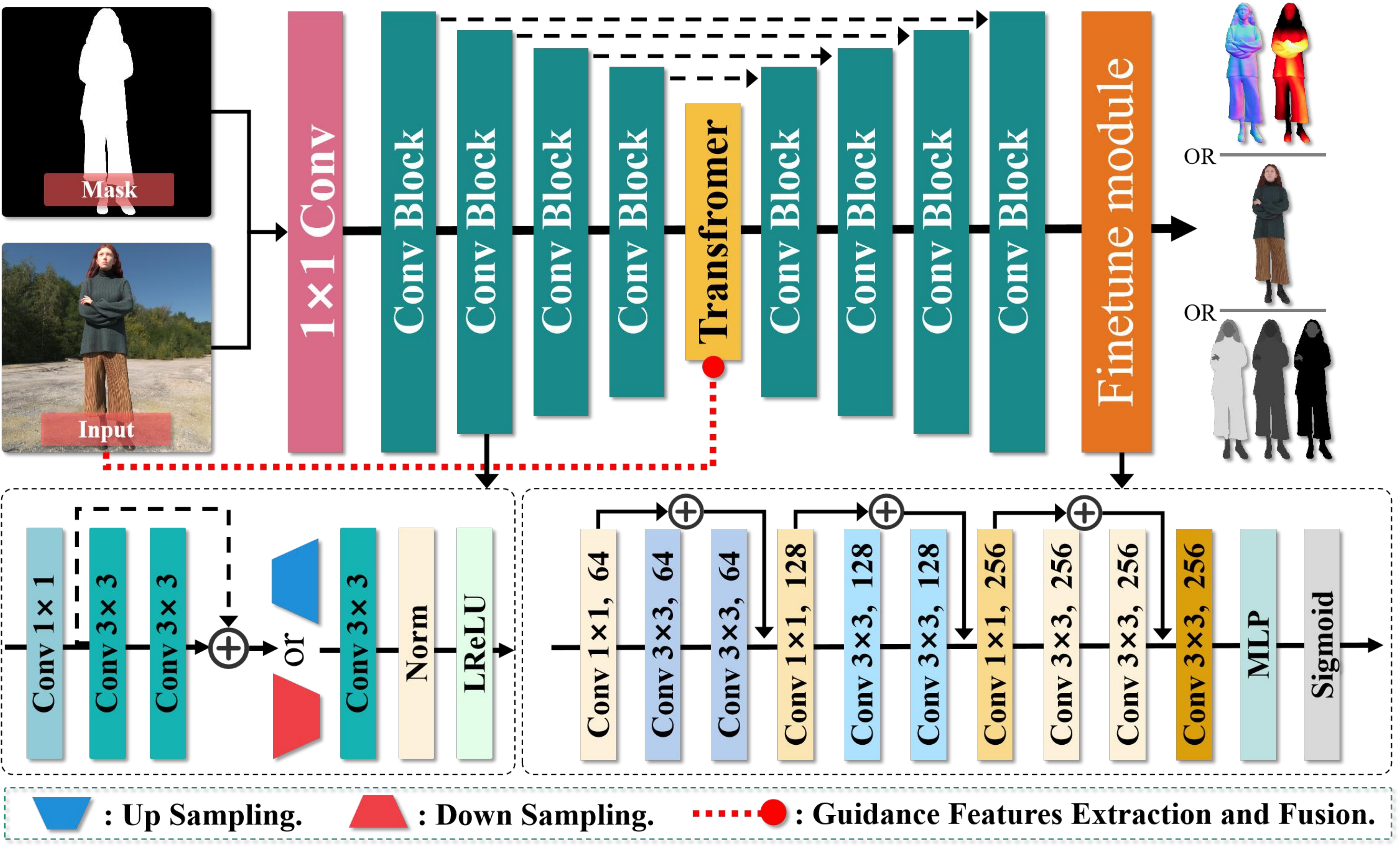}\vspace{\myvspace}   
			\end{minipage}
		}
	\end{minipage}
	\caption{
		The structure of the prior model. The network of the three prior models is similar, and their difference is only in the output material maps. Each convolutional block will perform up-sampling or down-sampling once. Similar to that in Fig.~\ref{overview}, we also use the Guidance Encoder to extract guidance features from the ``Input" and make fusion.
	}
	\label{img-prior_model}
\end{figure}
\Ours consists of two stages: (1) Take a single full-body image as input, employ \cite{Rembg} to extract the ``Foreground" (if the ``Mask" is not provided), and then employ three prior models to estimate materials priors (Sec.~\ref{prior_models}), which are trained on our \OursDataset dataset.
~(2) Based on the three pre-trained prior models, we design a guidance encoder for extracting guidance features from estimated prior materials. Subsequently, we design a \FinetuneModel (\FinetuneModelShort) to optimize the priors. This model takes the foreground map as input, and utilizes the guidance features to guide the decoding process.  After the joint optimization, the \FinetuneModelShort estimates the final results. 

\subsection{Prior Models}
\label{prior_models}
To achieve the goal of estimating six material maps from a single image and make it more practical, a direct and effective method is to use multiple models for material estimation. This can simplify the estimation process to some extent and improve accuracy and reliability. Moreover, training three models with different functions independently can make better use of resources when hardware is limited, compared to integrating all functions into one model.

According to the correlations between material maps, we carefully design the target estimated map(s) of each prior model. Specifically, normal and displacement information are closely related to the geometric surface. So, we create a \GeometryPriorModel~(\GeometryPriorModelShort) to estimate them.
~During data construction, we set the values of R, S, and SSS according to the material category. Based on this, we can effectively estimate them using only an \RSSPriorModel~(\RSSPriorModelShort). Notably, there is category-mixing at the boundaries of the material category. So, we cannot simply treat the estimation of R, S, and SSS as a classification task. 
~Last, we design an \AlbedoPriorModel~(\AlbedoPriorModelShort) to accurately estimate the remaining diffuse albedo.

\begin{figure}[!ht]
  \centering 
  \begin{minipage}[b]{\linewidth} 
  \centering 
  \newcommand{\myvspace}{1.0 pt} 
  \newcommand{\widthOfFullPage}{1} 
  \newcommand{\widthOfMiniPage}{1}
  \newcommand{\format}{png}
  \subfloat{
    \begin{minipage}[b]{\widthOfFullPage\linewidth} 
      \centering
    \includegraphics[width=\widthOfMiniPage\linewidth]{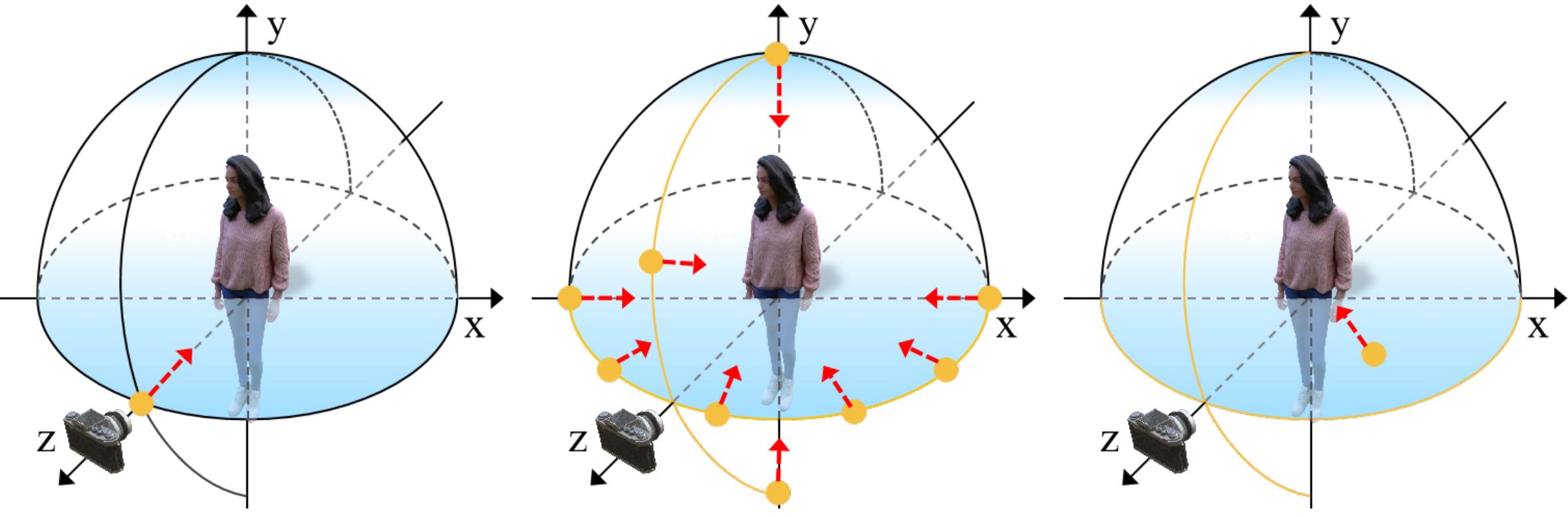}\vspace{\myvspace}
    \end{minipage}
  }
  \begin{flushleft}
  	\small
    \hspace{1.3em}(a) Fixed Single
    \hspace{2.8em}(b) Fixed Multi
    \hspace{3.2em}(c) Random
  \end{flushleft}
  \end{minipage} 
\caption{
Three kinds of light illuminations for PBR rendering. 
``Fixed Single" means single point light at a fixed position.
``Fixed Multi" means multi-point lights at fixed positions.
``Random" means single point light at a random position on positive hemisphere.
}
\label{img-rendering_illumination}
\end{figure}

As shown in Fig.~\ref{img-prior_model}, the three prior models have similarities in network structure. The input of each of them is ``Input" image and ``Mask", but the main difference among these three models lies in the channels of network output. Specifically, the output of the \GeometryPriorModelShort is four channels, which includes three channels of normal and one channel of displacement. The output of the \AlbedoPriorModelShort is three channels of diffuse albedo.  The output of the \RSSPriorModelShort is also three channels, which contains three single channels, namely roughness, specular albedo, and subsurface scattering.

The prior model presents an encoder-decoder structure. First, the input image undergoes a 1×1 convolution operation. Then, through 4 convolutional blocks for down-sampling, after extracting the guidance features from the ``Input", a transformer block is used to perform guidance features fusion and latent feature encoding, and then it goes through a four layer convolutional block for up-sampling. During the up-sampling process, features of the same scale are fused via the skip connections. Finally, the target image is output through a finetune module. The numbers of channels during down-sampling and up-sampling are 64, 128, 256, 512, 512, 512, 256, 128, and 64 in sequence. The structure of the convolutional block and the structure of the finetune module are shown in the lower left corner and lower right corner of Fig.~\ref{img-prior_model} respectively. The convolutional block contains 1×1 and 3×3 convolution operations, such as up-sampling and down-sampling, normalization, and activation functions. The finetune module contains three sets of convolution and residual connection operations. After passing through the multi-layer perceptron (MLP) and the sigmoid activation function, the target image can be obtained.
\begin{figure}[!b]
  \centering 
  \begin{minipage}[b]{\linewidth} 
  \centering 
  \newcommand{\myvspace}{1.0 pt} 
  \newcommand{\widthOfFullPage}{1} 
  \newcommand{\widthOfMiniPage}{0.98}
  \newcommand{\format}{png}
  \subfloat{
    \begin{minipage}[b]{\widthOfFullPage\linewidth} 
      \centering
    \includegraphics[width=\widthOfMiniPage\linewidth]{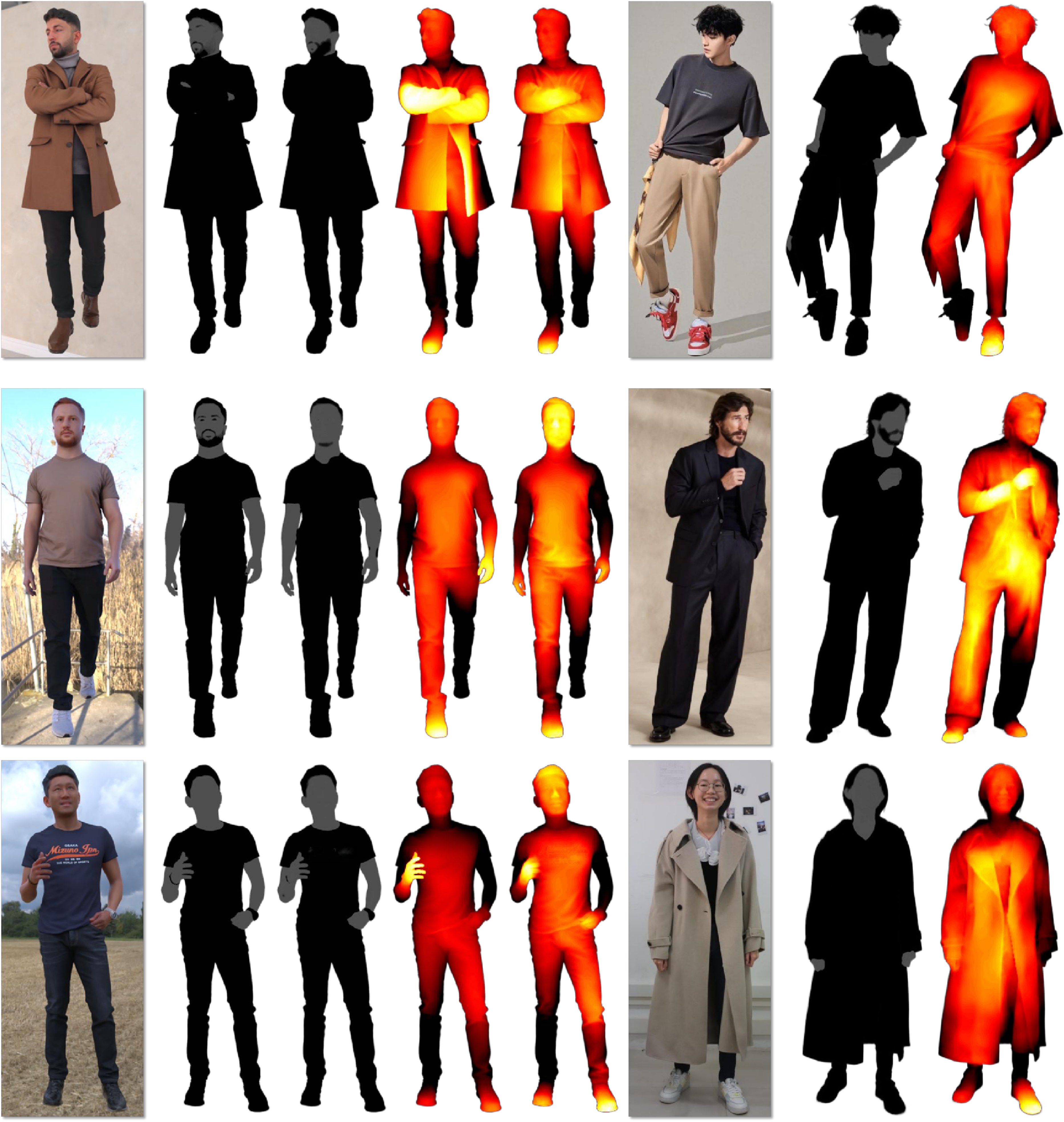}\vspace{\myvspace}   
    \end{minipage}
  }
  \begin{flushleft}
    \small
    \vspace{-0.5em}
    \hspace{1.7em}Input
    \hspace{1.7em}GT
    \hspace{1.2em}SSS
    \hspace{1.1em}GT
    \hspace{1.1em}Disp
    \hspace{1.1em}Input
    \hspace{1.6em}SSS
    \hspace{1.5em}Disp
  \end{flushleft}
  \end{minipage}
\caption{Input, estimated subsurface scattering, and displacement results. The ``Input" means the input image of Fig.~\ref{img-material_comparision_on_syn} and Fig.~\ref{img-material_comparision_on_real}. ``SSS" and ``Disp" mean the estimated subsurface scattering and displacement map of our method.
The left three samples are the estimated results on \OursDataset dataset, and the right three samples are the estimated result on the real data.
}
\label{img-material_comparision_supp}
\end{figure}

\subsection{\FinetuneModel}
\label{finetune_model}
Training three independent prior models to estimate multiple materials is an effective strategy, and it reduces the difficulty of this task which just uses a single model. But it also weakens the connection between various materials and makes it hard to guarantee the physical reasonability. 
The material priors estimated based on the prior model seem to have appropriate performance. Using it as the initial material and then further enhancing the rationality of the material combination through joint optimization is a reasonable finetuning strategy. Therefore, we construct a \FinetuneModel to handle this task.

As shown in Fig.~\ref{overview}, the \FinetuneModelShort contains a Latent Encoder, four material~(Nor, Disp, Diff, and RSS) decoders, and four Guidance Encoders. The input of \FinetuneModelShort is the ``Foreground" and the estimated priors of three prior models. The Latent Encoder encodes the ``Foreground" into latent features and serves as the input for the four material encoders. At the same time, the Guidance Encoders are used to encode the priors into guidance features and also serve as the input for the four material encoders. By fusing the latent features and the guidance features and then performing decoding, the final material can be obtained.

The structure of the Latent Encoder and four material decoders is the same as that of the prior models. The Guidance Encoder consists of nine convolutional blocks with channel numbers of 64, 64, 64, 128, 128, 128, 256, 256, and 256 in sequence. The structure of this convolutional block is consistent with the structure of the convolutional block in the prior model.
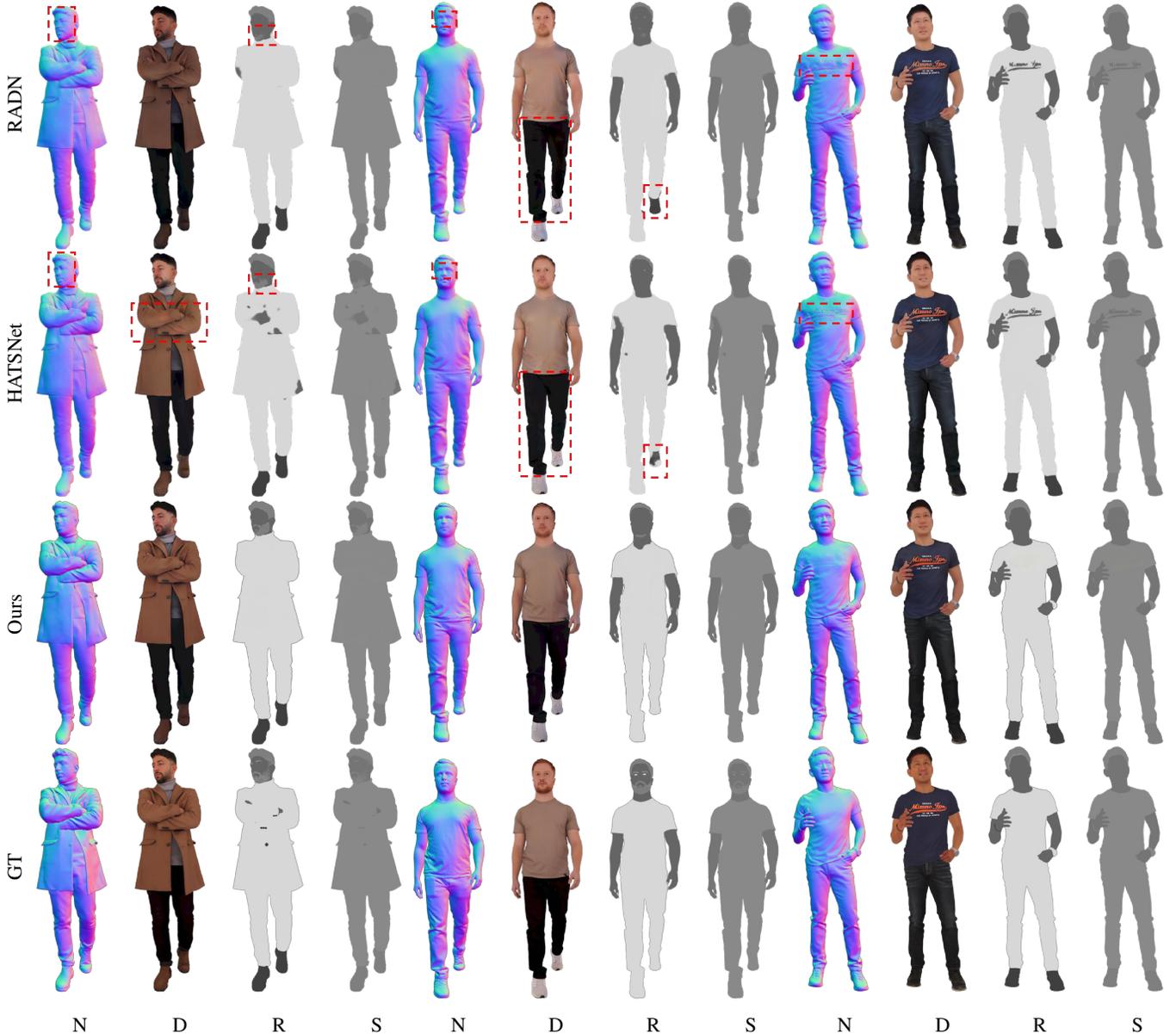
\begin{figure*}[!t]
  \centering 
  \begin{minipage}[b]{\linewidth} 
  \centering 
  \newcommand{\myvspace}{1.0 pt} 
  \newcommand{\widthOfFullPage}{1} 
  \newcommand{\widthOfMiniPage}{0.98}
  \newcommand{\format}{png}
  \subfloat{
    \begin{minipage}[b]{\widthOfFullPage\linewidth} 
      \centering
      \begin{picture}(0,0)
      	\small
        \put(0,60){\rotatebox{90}{\makebox(0,0)[c]{GT}}}
        \put(0,171){\rotatebox{90}{\makebox(0,0)[c]{Ours}}}
        \put(0,283){\rotatebox{90}{\makebox(0,0)[c]{\HATSNet}}}
        \put(0,397){\rotatebox{90}{\makebox(0,0)[c]{\RADN}}}
        \end{picture}
    \includegraphics[width=\widthOfMiniPage\linewidth]{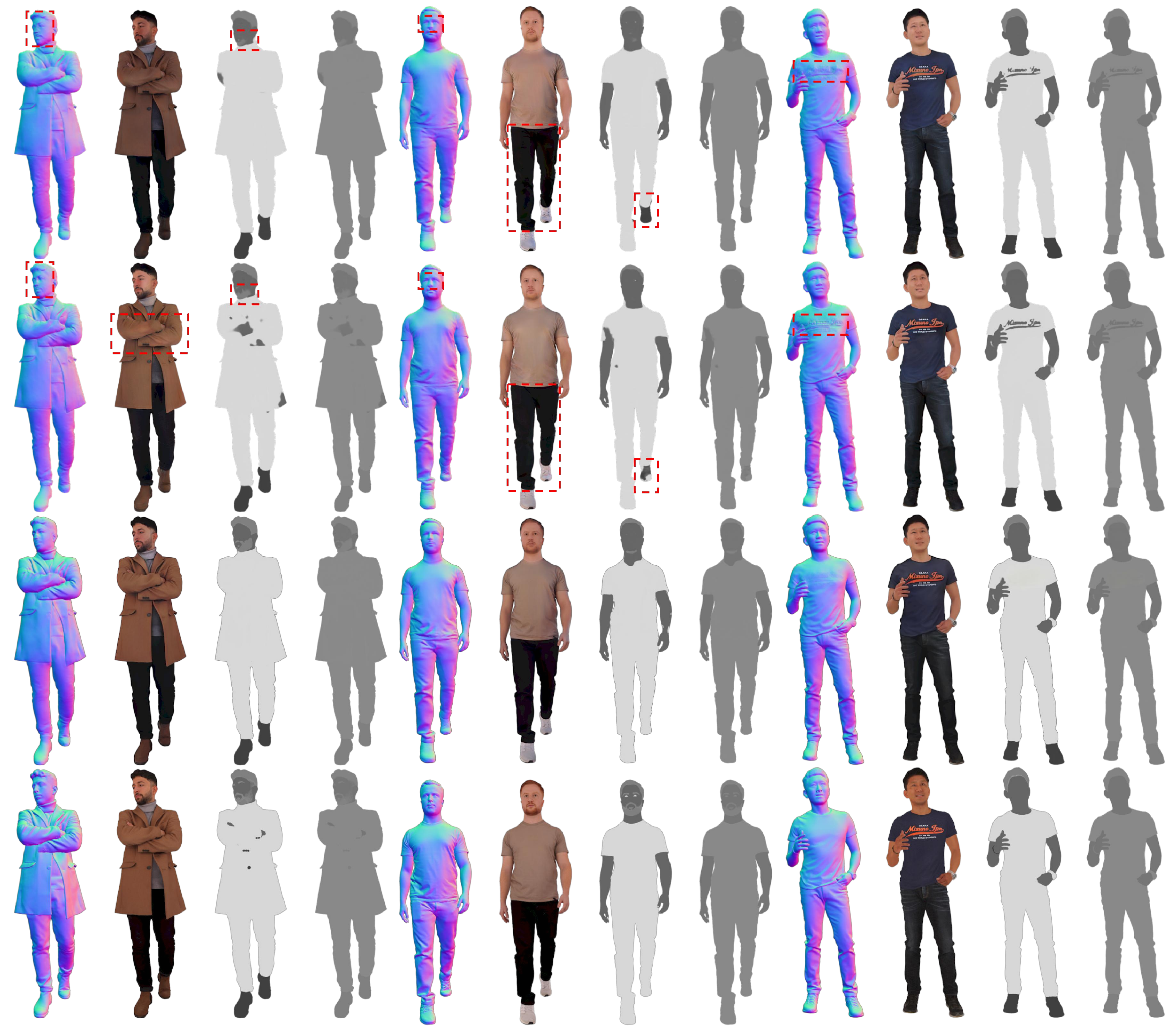}\vspace{\myvspace}   
    \end{minipage}
  }
  \begin{flushleft}
    \small
    \vspace{-0.5em}
    \hspace{3.6em}N
    \hspace{3.8em}D
    \hspace{3.8em}R
    \hspace{3.8em}S
    \hspace{3.0em}N
    \hspace{3.7em}D
    \hspace{3.7em}R
    \hspace{3.8em}S
    \hspace{3.6em}N
    \hspace{3.7em}D
    \hspace{3.7em}R
    \hspace{3.8em}S
  \end{flushleft}
  \end{minipage}
\caption{Performance comparison with previous works (\RADN~\cite{RADN_2018_TOG_single_image_natural_svbrdf}, and \HATSNet~\cite{HATSNet_2021_TOG_single_image_natural_svbrdf}) for PBR materials estimation on \OursDataset dataset. The ``Input", estimated ``Disp" and ``SSS" of ``Ours" are shown in Fig.~\ref{img-material_comparision_supp}. ``N" means normal, ``D" means diffuse albedo, ``R" means roughness, ``S" means specular albedo. The relighting results are shown in Fig.~\ref{img-relighting_comparision_of_svbrdf}. More details please see Sec.~\ref{performance_evaluation}. }
\label{img-material_comparision_on_syn}
\end{figure*}

\subsection{PBR Rendering}
\label{pbr_rendering}
For calculating rendering loss during the model training process, we employ a physicall-based rendering shader. This shader can effectively simulate the physical behavior of light rays. It is implemented based on the rendering equation, which considers the interaction and reflection of light rays on the surface.
~We calculate the surface reflection by:
\begin{equation}
	\label{reflection_equation}
	{L}(x) =  \int_{\Omega} L_{i} \cdot f_{r}(materials,{w}_{i},{w}_{o}) \cdot ({w}_{i} \cdot {n}) \cdot {dw}_{i},
\end{equation}
where $L$ represents the rendering result, $x$ is a point on the geometric surface, $L_{i}$ is light intensity. $materials$ consist of normal ($n$), diffuse albedo ($d$), roughness ($r$) , specular albedo ($s$), subsurface scattering ($sss$), and displacement ($disp$). 
~Note that the displacement represents the distance from the surface point $x$ under the current camera to the world center.
~$w_i$ and $w_o$ represent the incoming and outgoing directions of the light ray, respectively. $f_r$ is the BSDF, which quantitatively describes how surface reflectance changes with different directions of incident light, and we employ the Disney BSDF model~\cite{DisneyBRDF_2012_burley} to calculate it.
\begin{figure*}[!ht]
  \centering 
  \begin{minipage}[b]{\linewidth} 
  \centering 
  \newcommand{\myvspace}{1.0 pt} 
  \newcommand{\widthOfFullPage}{1} 
  \newcommand{\widthOfMiniPage}{0.985}
  \newcommand{\format}{png}
  \subfloat{
    \begin{minipage}[b]{\widthOfFullPage\linewidth} 
      \centering
      \begin{picture}(0,0)
      	\small
        \put(0,58){\rotatebox{90}{\makebox(0,0)[c]{Ours}}}
        \put(0,160){\rotatebox{90}{\makebox(0,0)[c]{\SwitchLight}}}
        \put(0,270){\rotatebox{90}{\makebox(0,0)[c]{\HATSNet}}}
        \put(0,381){\rotatebox{90}{\makebox(0,0)[c]{\RADN}}}
        \end{picture}
    \includegraphics[width=\widthOfMiniPage\linewidth]{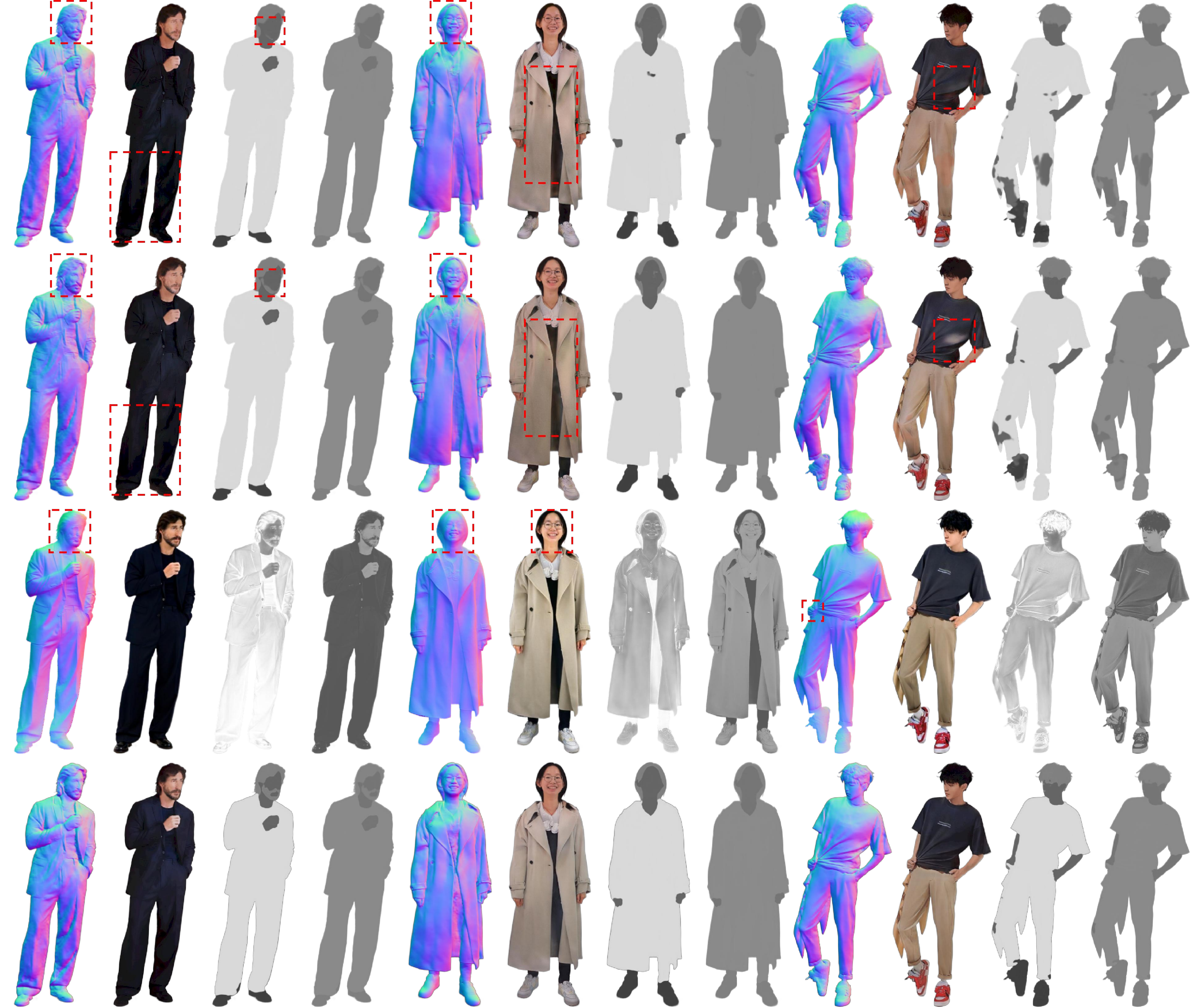}\vspace{\myvspace}   
    \end{minipage}
  }
  \begin{flushleft}
    \small
    \vspace{-0.5em}
    \hspace{2.9em}N
    \hspace{3.8em}D
    \hspace{3.5em}R
    \hspace{3.8em}S
	\hspace{3.9em}N
	\hspace{3.9em}D
	\hspace{3.6em}R
	\hspace{3.6em}S
	\hspace{3.2em}N
	\hspace{3.7em}D
	\hspace{3.7em}R
	\hspace{3.8em}S
  \end{flushleft}
  \end{minipage}
  
\caption{Performance comparison with previous works (\RADN~\cite{RADN_2018_TOG_single_image_natural_svbrdf}, \HATSNet~\cite{HATSNet_2021_TOG_single_image_natural_svbrdf}, and \SwitchLight~\cite{SwitchLight_2024_CVPR_pbr_and_neural_render}) for PBR materials estimation on real data. The ``Input", estimated ``Disp" and ``SSS" of ``Ours" are shown in Fig.~\ref{img-material_comparision_supp}. ``N" means normal, ``D" means diffuse albedo, ``R" means roughness, ``S" means specular albedo. The relighting results are shown in Fig.~\ref{img-relighting_comparision_of_svbrdf}. More details please see Sec.~\ref{performance_evaluation}.}
\label{img-material_comparision_on_real}
\end{figure*}
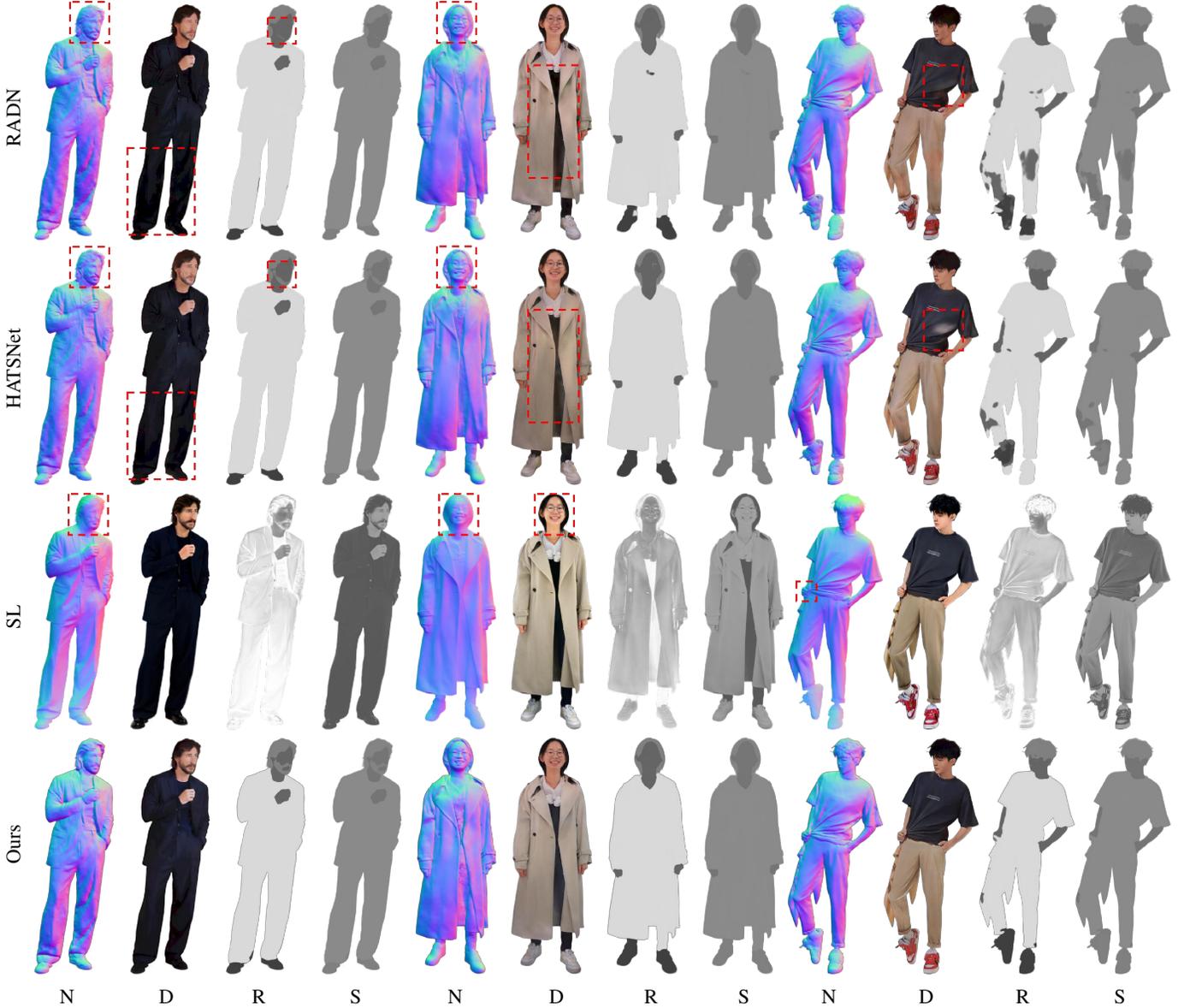

\subsection{Loss Function}
\label{loss_function}
Our model consists of three prior models and a finetuning model. During the training process, the losses of these models mainly include pixel-based L1 loss and rendering loss. Due to the different tasks of each model, we design a \ControlledPBRRendering (\ControlledPBRRenderingShort) loss for each model according to the estimated object.  Therefore, our method is trained over a joint loss function $\mathcal{L}_{total}$, which can be calculated by:
\begin{equation}
	\label{loss_total}
	\mathcal{L}_{total} = \underbrace{\mathcal{L}_{p}^g + \mathcal{L}_{cpr}^g}_{\text{Geometry}} + \underbrace{\mathcal{L}_{p}^a + \mathcal{L}_{cpr}^a}_{\text{Albedo}} + \underbrace{\mathcal{L}_{p}^r + \mathcal{L}_{r}^r}_{\text{RSS}} +  \underbrace{\mathcal{L}_{p}^m + \mathcal{L}_{r}^m}_{\text{Finetune}},
\end{equation}
where $\mathcal{L}_{p}$ means pixel-wised L1 loss, $\mathcal{L}_{cpr}$ means the \ControlledPBRRenderingShort loss, and $\mathcal{L}_{r}$ means the relighting loss. 

The pixel-wised L1 loss $\mathcal{L}_{p}$  can be calculated by:
\begin{equation}
	\label{loss_material}
	\mathcal{L}_{p} = \sum_{i}^{all}(||pred-gt||_1) ,
\end{equation}
where the $all$ depends on the estimated target of each model, for example, the targets of  \GeometryPriorModelShort are normal and displacement, while the target of \AlbedoPriorModelShort is diffuse albedo.

The single image material estimation is ill-posed, and an effective way to verify the accuracy of the material is to calculate the rendering loss in various illuminations. Thanks to our material data, we can produce the GT and the estimated material rendered in arbitrary illuminations during the training process. The more various illuminations for rendering, the more robust material estimation is.

Therefore, we design a multi-illuminations rendering loss for training. It contains $M$ (set to 37) kinds of illuminations for PBR rendering, where 36 are point lights under fixed and different positions (as shown in Fig.~\ref{img-rendering_illumination} (b)), and 1 is point light under random position (as shown in Fig.~\ref{img-rendering_illumination} (c)) with random light intensity (3.0-8.0). In Fig.~\ref{img-rendering_illumination} (b), within the \(XOZ\) and \(YOZ\) planes, the angular separation between adjacent point light sources is \(10^{\circ}\). This angular interval is sufficient to adequately depict the illumination direction in the majority of scenarios.  Based on GT and estimated materials, we render the appearance under those illuminations and calculate the loss, and the multi-illuminations rendering loss ($\mathcal{L}_{r}$) is calculated by:
\begin{equation}
	\label{loss_relighting}
	\begin{split}
		\mathcal L_{r}=\sum_{i}^{M}||(\mathcal{R}(gt, light_{i}), \mathcal{R}(pred, light_{i}))||_{1},
	\end{split}
\end{equation}
where $pred$ is estimated material, $gt$ is the GT material, $\mathcal{R}$ is the PBR shader,  $light_{i}$ is the illumination for rendering.

\begin{figure*}[!ht]
  \centering 
  \begin{minipage}[b]{\linewidth} 
  \centering 
  \newcommand{\myvspace}{1.0 pt} 
  \newcommand{\widthOfFullPage}{1} 
  \newcommand{\widthOfMiniPage}{1.0}
  \newcommand{\format}{png}
  \subfloat{
    \begin{minipage}[b]{\widthOfFullPage\linewidth} 
      \centering
    \includegraphics[width=\widthOfMiniPage\linewidth]{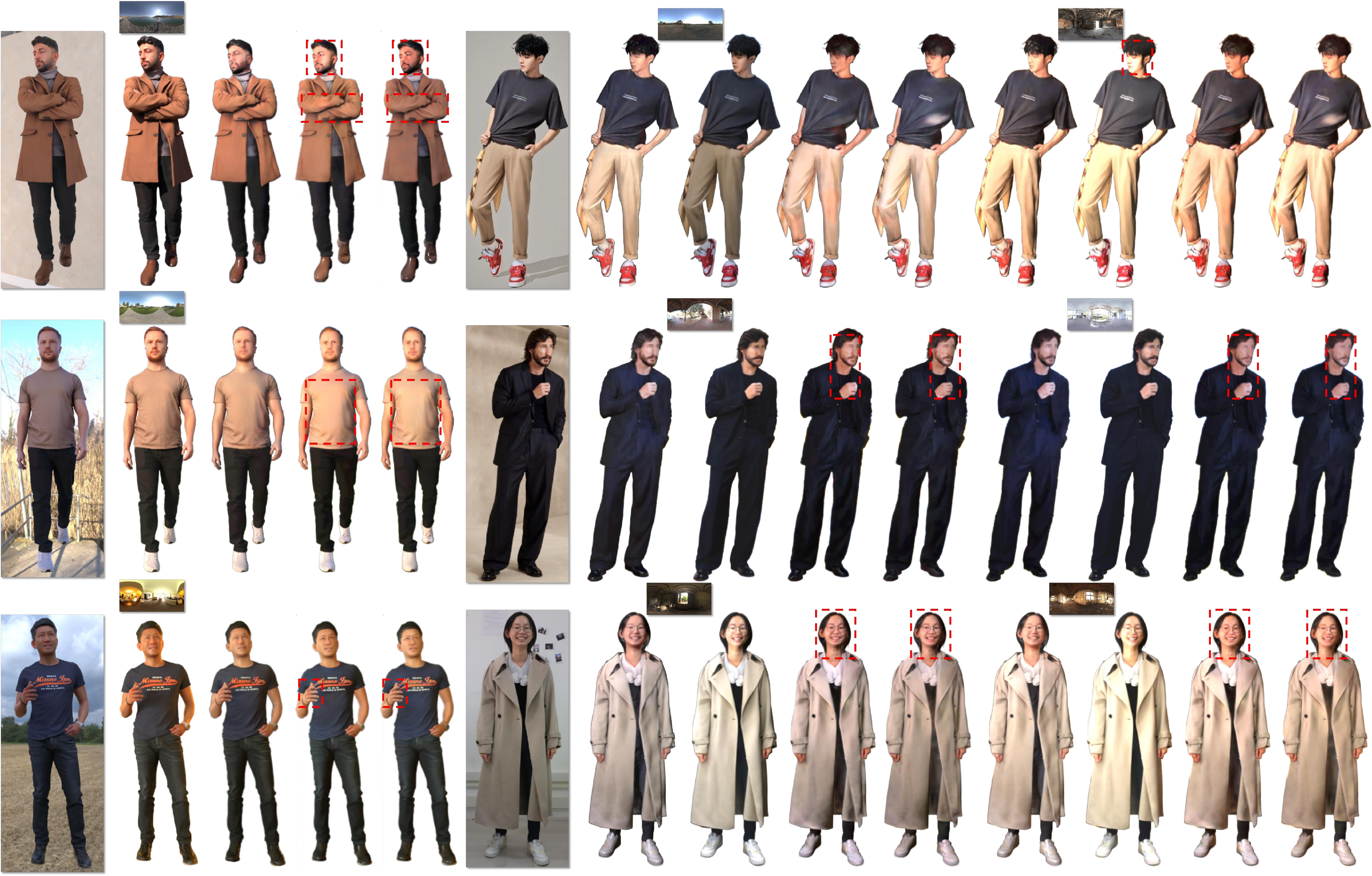}\vspace{\myvspace}   
    \end{minipage}
  }
  \begin{flushleft}
  	\small
  	\vspace{-0.5em}
  	\hspace{1.6em}Input
    \hspace{2.7em}GT
    \hspace{1.9em}Ours
    \hspace{0.1em}\HATSNet
    \hspace{-0.1em}\RADN
    \hspace{1.2em}Input
    \hspace{2.3em}Ours
    \hspace{2.3em}\SwitchLight
    \hspace{1.2em}\HATSNet
    \hspace{0.2em}\RADN
  	\hspace{1.7em}Ours
  	\hspace{2.3em}\SwitchLight
  	\hspace{1.2em}\HATSNet
  	\hspace{0.2em}\RADN
  \end{flushleft}
  \end{minipage}
\caption{Relighting results of the estimated materials.
~The estimated material maps are from Fig.~\ref{img-material_comparision_on_syn} (left samples are on the \OursDataset dataset) and Fig.~\ref{img-material_comparision_on_real} (right samples are results on the real data). Please see the rendering results under dynamic lighting in the attached video.}
\label{img-relighting_comparision_of_svbrdf}
\end{figure*}
The purpose of the \ControlledPBRRenderingShort loss function is to ensure that each material property can be optimized independently while maintaining physical rationality. The key idea is to fix the irrelevant parameters (either using GT data or setting them to physically reasonable values to minimize interference), so that the material to be optimized plays the most significant role during rendering process. In addition, when setting the irrelevant parameters, the realism of the material should be considered, and avoid setting extreme values to reduce the generalization ability of the model. 

Specifically, (1) for the \GeometryPriorModel, we need to make the highlights in the rendering results more likely to appear. This can enhance the impact of normal and displacement on the details of the rendering results. Therefore, we set the roughness to 0.2, and set a medium-intensity specular albedo (0.5).
(2) For the \AlbedoPriorModel, reducing occurrence of highlights can enhance the importance of diffuse albedo for the rendering results. Therefore, we set the roughness to 0.8 and the specular albedo to 0.03, minimizing the impact of specular reflection within the physically reasonable range. The GT data of subsurface scattering and normal are used to eliminate geometric interference. 
(3) For the \RSSPriorModel, the key of improving performance is to enhance the impact of the higlights on the rendering result. Based on this, we think that directly using GT data (normal, displacement, and diffuse albedo) can provide diversity in highlights, and it is the best option. 
In this way, the $\mathcal L_{cpr}$ can be defined by
\begin{equation}
	\label{loss_cpr}
	\begin{split}
		\mathcal L_{cpr}=\sum_{i}^{M}||(\mathcal{R}(pred,cm^{\prime},light_{i})-\mathcal{R}(gt,cm^{\prime},light_{i}))||_1,
	\end{split}
\end{equation}
where $\mathcal{R}$ is the PBR shader, $M$ (set to 37) means the kinds of illuminations for PBR rendering, $light_{i}$ is an illumination for rendering, and $cm^{\prime}$ means the controlled materials. It should be noted that HDR environment map is more complex than point light in terms of the number, intensity, and color of light sources. To avoid the influence of environment map on the rendering results exceeding the estimated materials, we only calculate the rendering loss under $M$ illuminations. The $M-1$ are point lights under fixed position, one is point light under random position with random light intensity.

\subsection{Implementation Details}
During finetuning model training, theoretically, the prior model needs to be loaded and the predicted material is used as the input of the material model. However, due to the limited graphic memory of the GPU, we cannot load the prior model and the material model simultaneously during training. Therefore, to make training feasible, when training the material model, we directly use the GT material as the input. When testing, the material predicted by the prior model is used as the input of the material model.

\begin{figure*}[!ht]
  \centering 
  \begin{minipage}[b]{\linewidth} 
  \centering 
  \newcommand{\myvspace}{1.0 pt} 
  \newcommand{\widthOfFullPage}{1} 
  \newcommand{\widthOfMiniPage}{1.0}
  \newcommand{\format}{png}
  \subfloat{
    \begin{minipage}[b]{\widthOfFullPage\linewidth} 
      \centering
    \includegraphics[width=\widthOfMiniPage\linewidth]{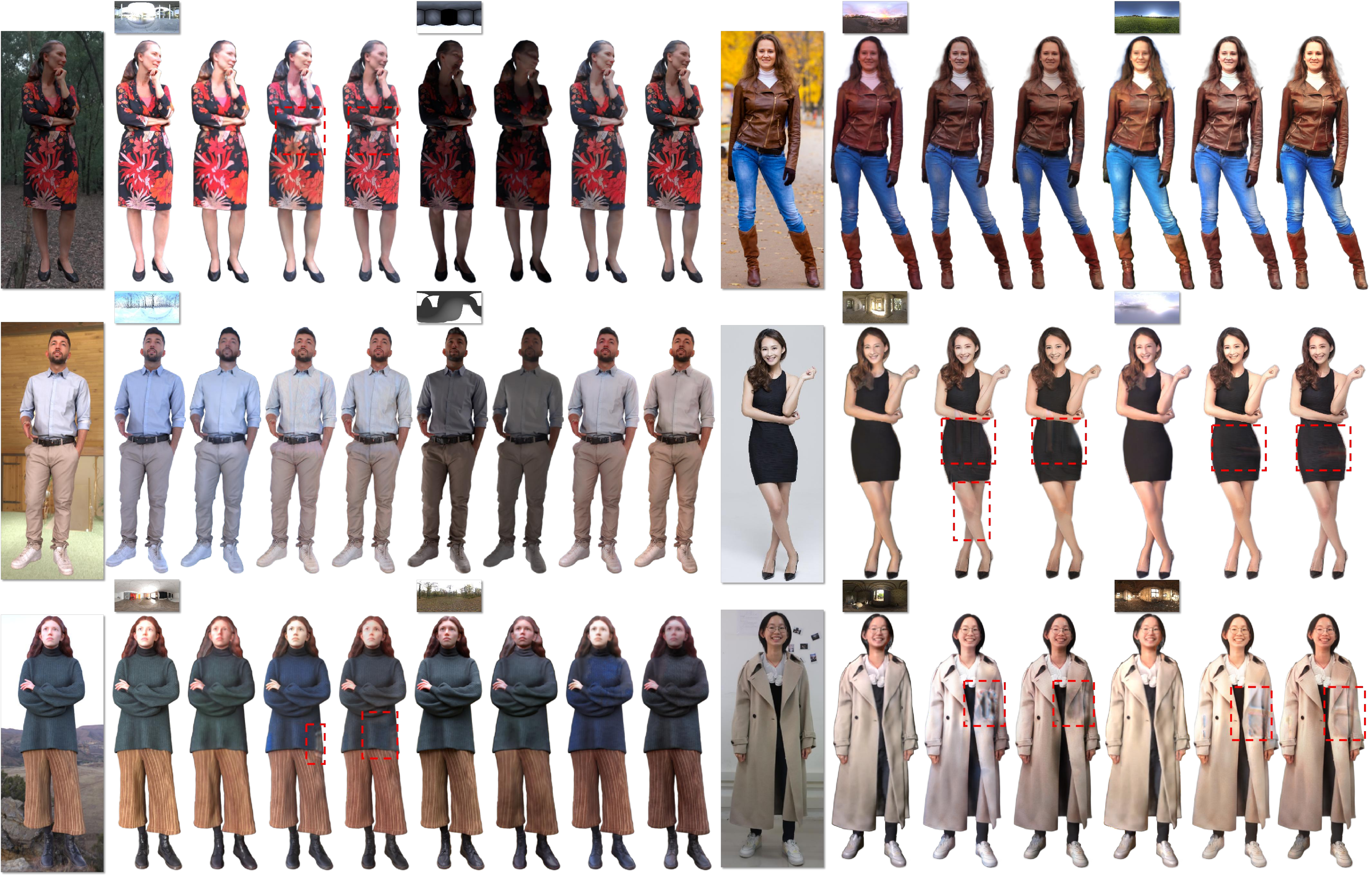}\vspace{\myvspace}   
    \end{minipage}
  }
  \begin{flushleft}
  	\small
  	\vspace{-0.5em}
  	\hspace{1.6em}Input
    \hspace{2.2em}GT
    \hspace{1.3em}Ours
    \hspace{0.6em}\FBHR
    \hspace{0.8em}\TotalReLighting
    \hspace{1.5em}GT
    \hspace{1.3em}Ours
    \hspace{0.6em}\FBHR
    \hspace{0.7em}\TotalReLighting
    \hspace{1.8em}Input
    \hspace{1.7em}Ours
    \hspace{1.3em}\FBHR
    \hspace{1.5em}\TotalReLighting
    \hspace{1.7em}Ours
    \hspace{1.1em}\FBHR
    \hspace{1.3em}\TotalReLighting
  \end{flushleft}
  \end{minipage}
\caption{
Performance comparison with previous works (\FBHR~\cite{Lagunas_2021_EGSR_single_image_human_neural_render}, \TotalReLighting~\cite{TotalRelighting_2021_Sig_single_image_human_phong_neural_render}) for relighting on \OursDataset dataset and real data.
~The left three are the results on \OursDataset dataset, and the right three are the results on real data.
~More details see Sec.~\ref{performance_evaluation}.}
\label{img-relighting_comparision}
\end{figure*}
Three prior models and one material model are trained independently. All these models are trained for up to 100 epochs with a batch size of 1 on single NVIDIA RTX 3090 GPU with 24GB of memory. Completing such a training process takes about two day for each Prior Model and three day for \FinetuneModel.
We implement our method using TensorFlow~\cite{Tensorflow_2016_Abadi}. We employ the AdamW optimizer for loss optimization, and the learning rate is set to 1e-4.

\begin{table*}
	\centering
	\caption{Materials and rendering result error (in terms of PSNR) comparison between previous works (\RADN\cite{RADN_2018_TOG_single_image_natural_svbrdf}, \HATSNet\cite{HATSNet_2021_TOG_single_image_natural_svbrdf}, \FBHR\cite{Lagunas_2021_EGSR_single_image_human_neural_render}, and \TotalReLighting\cite{TotalRelighting_2021_Sig_single_image_human_phong_neural_render}) on \OursDataset dataset. ``N" means normal, ``D" means diffuse albedo, ``R" means roughness, ``S" means specular albedo, ``SSS" means subsurface scatting. ``Disp" means displacement. ``Point" means relighting under the fixed single point light as shown in Fig.~\ref{img-rendering_illumination}. ``Real" means relighting under four real HDR environment maps. ``Syn" means relighting under a synthetic HDR environment map. 'Relit' means the relighting result. Best scores are highlighted in bold, and the second scores are underlined.}
	\label{Table-comparison_on_openhumanbrdf_dataset}
	\begin{tblr}{
			row{1} = {c},
			cell{1}{1} = {r=3}{c},
			cell{1}{2} = {c=6}{c},
			cell{1}{8} = {c=6}{c},
			cell{1}{14} = {c=3}{c},
			cell{2}{2} = {r=2}{c},
			cell{2}{3} = {r=2}{c},
			cell{2}{4} = {r=2}{c},
			cell{2}{5} = {r=2}{c},
			cell{2}{6} = {r=2}{c},
			cell{2}{7} = {r=2}{c},
			cell{2}{8} = {r=2}{c},
			cell{2}{9} = {c=4}{c},
			cell{2}{13} = {r=2}{c},
			cell{2}{14} = {r=2}{c},
			cell{2}{15} = {r=2}{c},
			cell{2}{16} = {r=2}{c},
			cell{4}{2} = {c},
			cell{4}{3} = {c},
			cell{4}{4} = {c},
			cell{4}{5} = {c},
			cell{4}{6} = {c},
			cell{4}{7} = {c},
			cell{4}{8} = {c},
			cell{4}{9} = {c},
			cell{4}{14} = {c},
			cell{4}{16} = {c},
			cell{5}{2} = {c},
			cell{5}{3} = {c},
			cell{5}{4} = {c},
			cell{5}{5} = {c},
			cell{5}{6} = {c},
			cell{5}{7} = {c},
			cell{5}{8} = {c},
			cell{5}{9} = {c},
			cell{5}{14} = {c},
			cell{5}{16} = {c},
			cell{6}{2} = {c},
			cell{6}{3} = {c},
			cell{6}{4} = {c},
			cell{6}{5} = {c},
			cell{6}{6} = {c},
			cell{6}{7} = {c},
			cell{6}{8} = {c},
			cell{6}{9} = {c},
			cell{6}{14} = {c},
			cell{6}{16} = {c},
			cell{7}{2} = {c},
			cell{7}{3} = {c},
			cell{7}{4} = {c},
			cell{7}{5} = {c},
			cell{7}{6} = {c},
			cell{7}{7} = {c},
			cell{7}{8} = {c},
			cell{7}{9} = {c},
			cell{7}{14} = {c},
			cell{7}{16} = {c},
			cell{8}{2} = {c},
			cell{8}{3} = {c},
			cell{8}{4} = {c},
			cell{8}{5} = {c},
			cell{8}{6} = {c},
			cell{8}{7} = {c},
			cell{8}{8} = {c},
			cell{8}{9} = {c},
			cell{8}{14} = {c},
			cell{8}{16} = {c},
			hline{1,9} = {-}{0.13em},
			hline{2} = {2}{l}, hline{2} = {3-6}{},hline{2} = {7}{r},
			hline{2} = {8}{l},hline{2} = {9-12}{},hline{2} = {13}{r},
			hline{2} = {14}{l},hline{2} = {15-16}{},hline{2} = {17}{r},
			hline{3} = {9}{l}, hline{3} = {10-11}{}, hline{3} = {12}{r},
			hline{4} = {-}{0.1em},
		}
		Method                                        & Material &   &   &   &     &      & Relighting &         &         &         &         &     & Mean      &            &       \\
		& N         & D & R & S & SSS & Disp & Point     & Real    &         &         &         & Syn & Material & Relit.& Total \\
		&           &   &   &   &     &      &           & Real\_1 & Real\_2 & Real\_3 & Real\_4 &     &           &            &       \\
		\RADN              &20.3    &26.0   &22.4   &38.1   & /      & /      &21.4   &21.5     &21.9         &21.7       &21.8      &22.6     &26.7  &21.9   &24.3 \\
		\HATSNet         &\underline{20.5}    &\underline{26.2}  &\underline{22.9}   &38.4   & /      & /      &\underline{21.9}   &21.7     &21.9         &21.8        &22.0     &\underline{22.8}     &\underline{27.0}   &22.0  &\underline{24.5} \\
		\FBHR               & /         &25.7   & /        & /        & /      & /      & /        &21.9     &22.7         &\underline{22.4}        &\underline{22.1}     &20.5     &25.7   &21.9  &23.8 \\
		\TotalReLighting &19.5    &25.9   & /        & /        & /      & /      & /        &\underline{22.0}     &\textbf{23.1}         &\textbf{22.6}        &\textbf{22.3}     &20.7     &22.7   &\underline{22.1}  &22.4 \\
		Ours                  &\textbf{21.2}    &\textbf{27.1}   &\textbf{24.1}   &\textbf{39.5}   &\textbf{41.6}&\textbf{28.8} &\textbf{23.2}   &\textbf{22.4}     &\underline{22.9}         &\underline{22.5}        &\textbf{22.3}     &\textbf{24.5}     &\textbf{30.4}   &\textbf{22.9}   &\textbf{26.7}       
	\end{tblr}
\end{table*}

\begin{table*}
	\centering
	\caption{Materials and rendering result error (in terms of PSNR) of ablation study on \OursDataset dataset.}
	\label{table-ablation}
	\begin{tblr}{
			row{1} = {c},
			cell{1}{1} = {r=3}{c},
			cell{1}{2} = {c=6}{c},
			cell{1}{8} = {c=6}{c},
			cell{1}{14} = {c=3}{c},
			cell{2}{2} = {r=2}{c}, 
			cell{2}{3} = {r=2}{c},
			cell{2}{4} = {r=2}{c},
			cell{2}{5} = {r=2}{c},
			cell{2}{6} = {r=2}{c},
			cell{2}{7} = {r=2}{c},
			cell{2}{8} = {r=2}{c},
			cell{2}{9} = {c=4}{c},
			cell{2}{13} = {r=2}{c},
			cell{2}{14} = {r=2}{c},
			cell{2}{15} = {r=2}{c},
			cell{2}{16} = {r=2}{c},
			cell{3-10}{2-16} = {c},
			hline{1,11} = {-}{0.13em},
			hline{2} = {2}{l}, hline{2} = {3-6}{},hline{2} = {7}{r},
			hline{2} = {8}{l},hline{2} = {9-12}{},hline{2} = {13}{r},
			hline{2} = {14}{l},hline{2} = {15-16}{},hline{2} = {17}{r},
			hline{3} = {9}{l}, hline{3} = {10-11}{}, hline{3} = {12}{r},
			hline{4} = {-}{0.1em},
		}
		Method    & Material &   &   &   &     &      & Relighting &         &         &         &         &     & Mean      &            &       \\
		& N         & D & R & S & SSS & Disp & Point     & Real    &         &         &         & Syn & Material & Relit.& Total \\
		&           &   &   &   &     &      &           & Real\_1 & Real\_2 & Real\_3 & Real\_4 &     &           &            &       \\
		w/o Guid.     &20.2 &25.7   &22.4   &37.5   &32.4  &27.3 &22.0  &21.5  &21.7  &21.6  &21.5  &23.1 &27.6  &21.9 &24.7  \\
		w/o P-Guid. &20.0 &26.4   &22.2   &37.9   &32.3  &27.7 &22.4  &21.8  &22.2  &22.1  &21.9  &23.3 &27.8  &22.3 &25.0  \\
		w/o Opt-C.  &20.5 &26.4   &\underline{24.0}   &\underline{39.2}   &\underline{40.3}  &28.6 &22.8  &21.8  &22.1  &21.9  &21.8  &23.2 &29.8  &22.2 &26.0  \\
		w/o Opt.      &\underline{20.8} &\underline{26.8}   &\underline{24.0}   &\underline{39.2}   &\underline{40.3}  &\underline{28.7} &\underline{23.0}  &\underline{21.9}  &22.1  &22.0  &\underline{22.0}  &\underline{23.4} &\underline{30.0}  &\underline{22.3} &\underline{26.1}  \\
		w/o SSS      & /      & /        & /         & /        & /       & /      &22.7  &\underline{21.9}  &22.2  &22.0  &21.9  &22.9  & /       &22.2 &22.2 \\
		w/o Disp.     & /      & /        & /         & /        & /       & /      &22.2  &21.7  &\underline{22.4}  &\underline{22.1}  &21.8  &23.2  & /       &22.1 &22.1 \\
		Full              &\textbf{21.2}    &\textbf{27.1}   &\textbf{24.1}   &\textbf{39.5}   &\textbf{41.6}&\textbf{28.8} &\textbf{23.2}   &\textbf{22.4}     &\textbf{22.9}         &\textbf{22.5}        &\textbf{22.3}     &\textbf{24.5}     &\textbf{30.4}   &\textbf{22.9}   &\textbf{26.7}      
	\end{tblr}
\end{table*}
\section{Experiments}
\label{experiment}
We first conduct qualitative and quantitative experiments to validate the performance of our method on \OursDataset dataset and real data. Then, we conduct ablation study to verify the effectiveness of training strategies and each component.

\subsection{Comparison methods}
To validate the effectiveness of our method, we compare it with related works in two aspects: PBR material estimation, and relighting.

\textbf{PBR material estimation.} ~\RADN\cite{RADN_2018_TOG_single_image_natural_svbrdf} and ~\HATSNet\cite{HATSNet_2021_TOG_single_image_natural_svbrdf} estimate materials from a near-planar image, both methods can estimate high-quality PBR materials, including normal, diffuse albedo, roughness, and specular albedo. We have trained them on \OursDataset from scratch.
~\SwitchLight\cite{SwitchLight_2024_CVPR_pbr_and_neural_render} has similar goals to our work, and performs human relighting by combining a physics-guided architecture with a pre-training framework. With a self-supervised pre-training strategy, ~\SwitchLight\ can estimate PBR materials, including normal, diffuse albedo, roughness, and specular albedo. Although the authors do not release their dataset and code, they provide a website for inference. For fair comparison, we only compare our method with \SwitchLight on the real data. 

\textbf{Relighting.}
~We compare the performance of relighting with \FBHR~\cite{Lagunas_2021_EGSR_single_image_human_neural_render} and \TotalReLighting~\cite{TotalRelighting_2021_Sig_single_image_human_phong_neural_render}. 
~\FBHR decomposes human appearance into reflectance and a light-dependent residual term, achieving impressive relighting effects even though it only estimates single material (diffuse albedo). 
~\TotalReLighting introduces a per-pixel lighting representation in a deep learning framework and explicitly models the diffuse and the specular components of appearance, producing convincing relighting portraits.
Based on our dataset, we can produce the training data required for \FBHR and \TotalReLighting, and we have trained them from scratch.

\subsection{Metrics}
To quantitatively evaluate the material estimation performance of our method and the comparative methods, we calculate the PSNR between the material maps estimated by the model and the GT data (Table~\ref{Table-comparison_on_openhumanbrdf_dataset}). In addition, we also calculate the PSNR of rendering results under various illuminations, such as point lights, real environment maps, and synthetic environment maps.

\subsection{Performance Evaluation}
\label{performance_evaluation}
To demonstrate the effectiveness of our method (\Ours), we conduct comparative experiments on \OursDataset dataset and real data. The input images for material estimation are shown in Fig.~\ref{img-material_comparision_supp}.
\begin{figure}[!b]
  \centering 
  \begin{minipage}[b]{\linewidth} 
  \centering 
  \newcommand{\myvspace}{1.0 pt} 
  \newcommand{\widthOfFullPage}{1} 
  \newcommand{\widthOfMiniPage}{0.92}
  \newcommand{\format}{png}
  \subfloat{
    \begin{minipage}[b]{\widthOfFullPage\linewidth} 
        \centering
        \includegraphics[width=\widthOfMiniPage\linewidth]{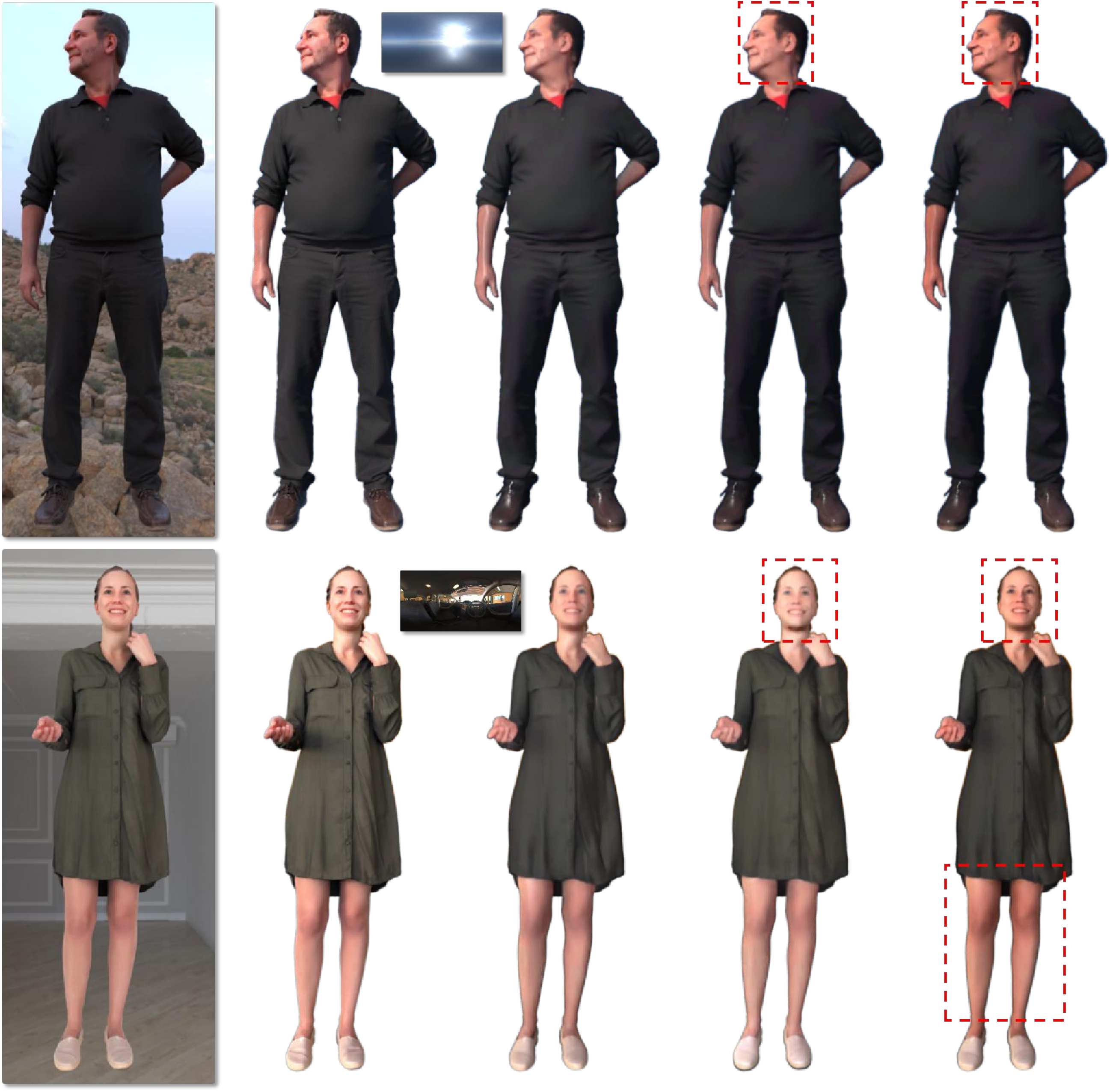}
        \begin{flushleft}
        	\small
        	\vspace{-0.5em}
        	\hspace{2.8em}Input
        	\hspace{3.0em}GT
        	\hspace{3.3em}Full
        	\hspace{2.1em}w/o Disp
        	\hspace{1.1em}w/o SSS
        \end{flushleft}
    \end{minipage}
  }
  \end{minipage}
\caption{Ablation study for realism enhancement on \OursDataset dataset.  ``w/o SSS." means not using the subsurface scattering during relighting. ``w/o Disp." means not using the displacement during relighting and approximate the surface geometry as a plane.
~More details please see Sec.~\ref{ablation_study_on_openhumanbrdf_datset}.}
\label{img-ablation_on_openhumanbrd_sup}
\end{figure}

As shown in Table~\ref{Table-comparison_on_openhumanbrdf_dataset}, our method almost achieves the best and competitive performance in both ``Material" and ``Relighting". The main reason for this is that the progressive training strategy of our method can reduce the interference of various materials during the optimization process, making it easier to achieve model fitting. The more accurate the estimated material is, the higher the performance of the relighting result will be.
The relighting methods under real illumination based on neural shader methods (\FBHR and \TotalReLighting) have better ``Relighting" performance than those based on PBR methods (\RADN and \HATSNet), but this result was not reproduced under synthetic illumination. Our results are competitive with the neural shader method in terms of ``Relighting" performance under ``Real", but our method outperforms under ``Syn".
This is because the neural shader methods rely on the kinds of illumination in the training data, and the relighting effect is prone to overfitting. When there are no new or similar illuminations for relighting in the training data, these methods struggle to achieve accurate relighting. The PBR-based methods can estimate render-ready materials and can perform robust rendering under arbitrary illuminations.
\begin{figure}[!b]
  \centering 
  \begin{minipage}[b]{\linewidth} 
  \centering 
  \newcommand{\myvspace}{1.0 pt} 
  \newcommand{\widthOfFullPage}{1} 
  \newcommand{\widthOfMiniPage}{0.98}
  \newcommand{\format}{png}
  \subfloat{
    \begin{minipage}[b]{\widthOfFullPage\linewidth} 
        \centering
        \includegraphics[width=\widthOfMiniPage\linewidth]{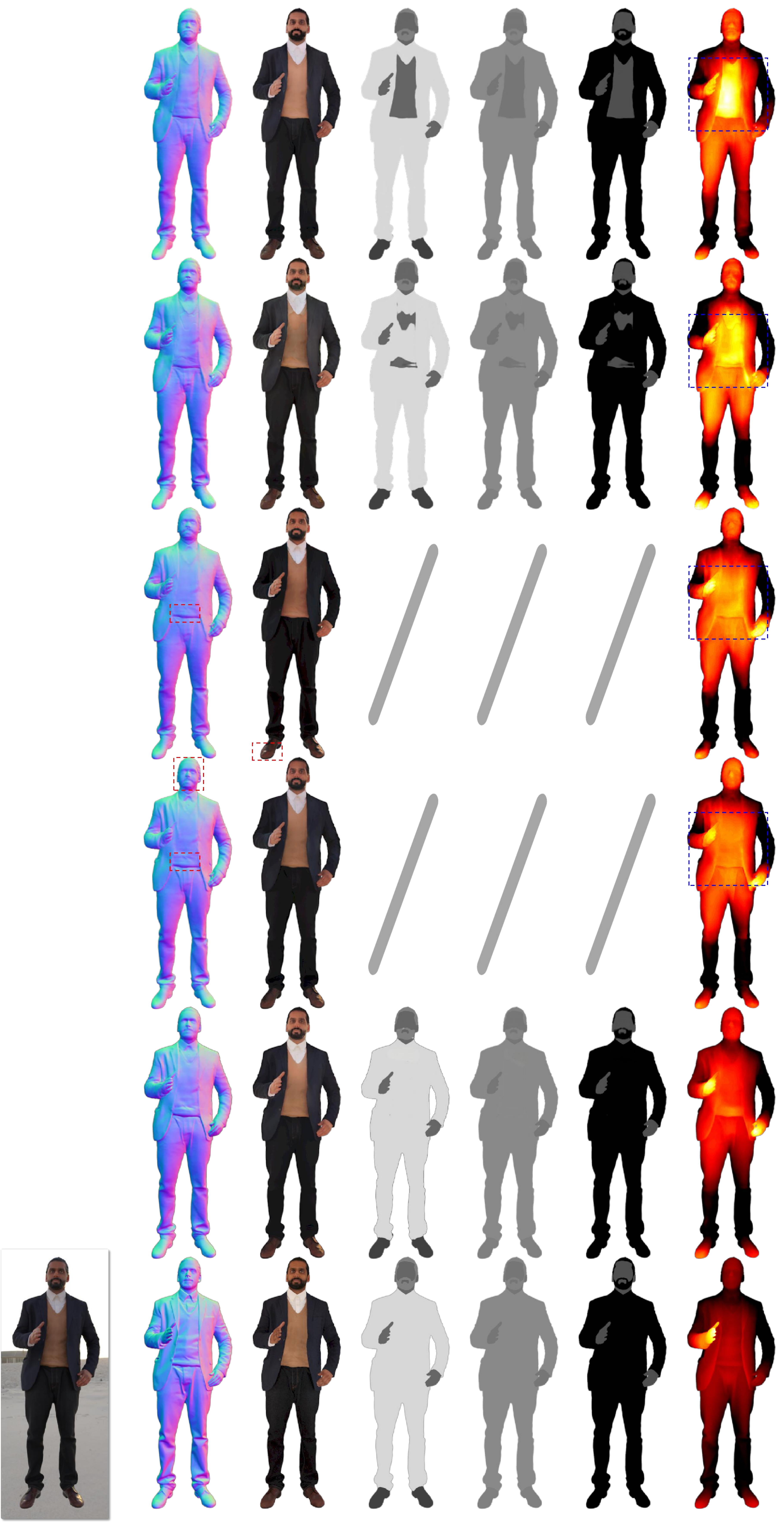}
        \begin{picture}(0,0)
        	\small
            \put(-211,46){\rotatebox{90}{\makebox(0,0)[c]{GT}}}
            \put(-211,125){\rotatebox{90}{\makebox(0,0)[c]{Full}}}
            \put(-211,202){\rotatebox{90}{\makebox(0,0)[c]{w/o Opt}}}
            \put(-211,280){\rotatebox{90}{\makebox(0,0)[c]{w/o Opt-C}}}
            \put(-211,360){\rotatebox{90}{\makebox(0,0)[c]{w/o P-Guid}}}
            \put(-211,439){\rotatebox{90}{\makebox(0,0)[c]{w/o Guid}}}
        \end{picture}
    \end{minipage}
  }
  \begin{flushleft}
  	\small
  	\vspace{-0.5em}
    \hspace{1.4em}Input
    \hspace{2.8em}N
    \hspace{2.8em}D
    \hspace{2.8em}R
    \hspace{2.8em}S
    \hspace{2.4em}SSS
    \hspace{1.6em}Disp
  \end{flushleft}
  \end{minipage}
\caption{Ablation study on \OursDataset dataset. 
~``N", ``D", ``R", ``S", ``SSS", and ``Disp" represent the normal, diffuse albedo, roughness, specular albedo, subsurface scattering, and displacement respectively. 
~More details see Sec.~\ref{ablation_study_on_openhumanbrdf_datset}.}
\label{img-ablation_on_openhumanbrd}
\end{figure}
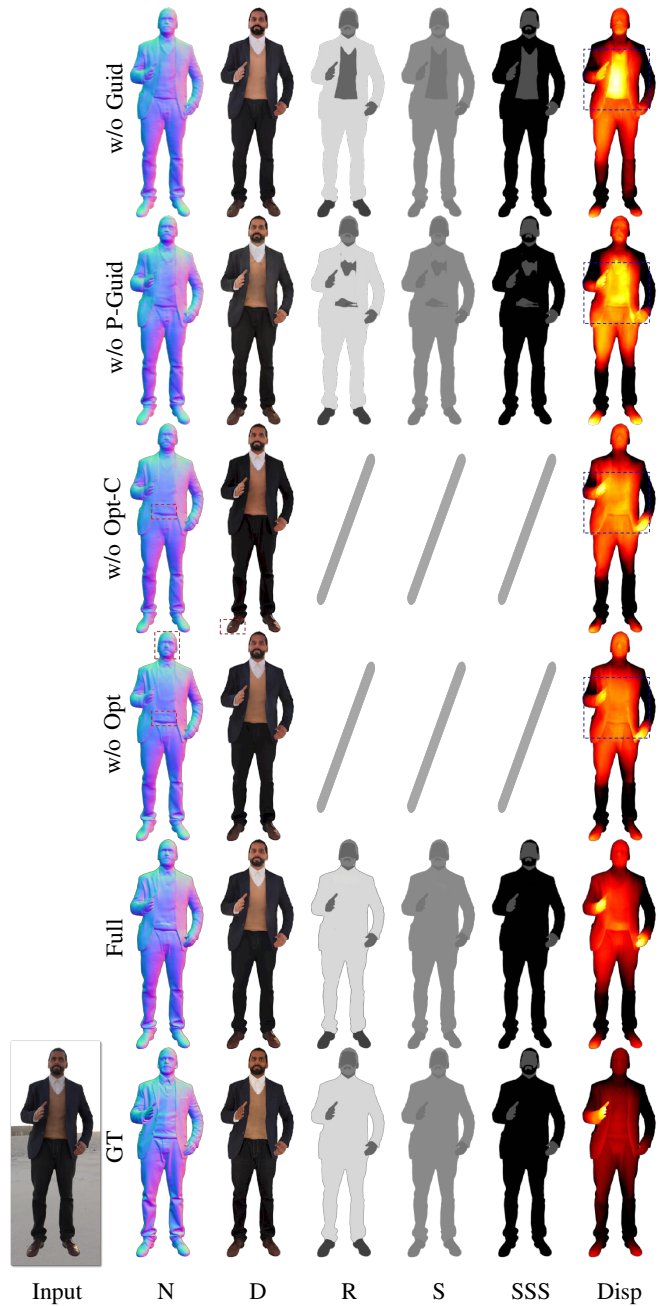

As shown in Fig.~\ref{img-material_comparision_on_syn} and Fig.~\ref{img-material_comparision_on_real}, thanks to the progressive training strategy, compared with the comparison methods, the normal we estimated has more details, and the hue of diffuse albedo is more accurate.
\SwitchLight have not trained on our dataset, the shadows or highlights are baked in the estimated diffuse albedo (Fig.~\ref{img-material_comparision_on_real}), and the baked shadow of other methods trained on our dataset is less. It proves our dataset is close to the real scene, and it is valid to estimate higher-quality diffuse albedo.
~Besides, we show the relighting results of the materials estimated in Fig.~\ref{img-relighting_comparision_of_svbrdf}. Compared to other methods, our method can estimate the subsurface scattering (Fig.~\ref{img-material_comparision_supp}), which helps to produce the realistic highlights in the skin areas.

To evaluate the relighting performance with neural shader method, we conduct comparison experiments with \FBHR and \TotalReLighting on \OursDataset dataset and real data (Fig.~\ref{img-relighting_comparision}). 
When the novel illumination is complex, the neural render-based method struggles to achieve accurate relighting results.
~Complex lighting scenes are mainly manifested as: (1) the lighting of input image is different from the novel illumination; (2) the difference (color or intensity) between the light sources in the environment map is large.
~Under complex novel illuminations, \FBHR and \TotalReLighting produce some artifacts and are hard to make realistic relighting results. Thanks to the estimated accurate render-ready PBR material maps, our method achieves accurate relightings under complex illuminations and produces more realistic skin reflection.

\subsection{Ablation Study}
\label{ablation_study_on_openhumanbrdf_datset}
We conduct extensive ablation experiments on \OursDataset dataset and real data to verify the effectiveness of our training strategy and each component. Specifically, we illustrate the effects of finetuning, \ControlledPBRRenderingShort loss, guidance feature fusion, multi-illuminations rendering, effects of  subsurface scattering, and effects of displacement with six different settings:
(1) ``w/o Guid" means that the finetuning model does not use guidance map for decoding but directly uses the \FinetuneModel to estimate final material maps.
(2) ``w/o P-Guid" means that the finetuning model uses input image as the guidance map and directly uses the \FinetuneModel to estimate final material maps.
(3) ``w/o Opt-C" means that the \ControlledPBRRenderingShort loss is not used in the process of prior model training and directly estimates the material  results by the prior models.
(4) ``w/o Opt" means directly using the material estimation results of the prior models without finetuning.
(5) ``w/o SSS" means not using the subsurface scattering during relighting.
(6) ``w/o Disp" means not using the displacement during relighting, which means approximating the surface geometry as a plane.

As shown in Table~\ref{Table-comparison_on_openhumanbrdf_dataset} and Fig.~\ref{img-ablation_on_openhumanbrd}, through careful observation, we find the results of full model (``Full") are closest to the GT. 
Based on well-designed structure, ``w/o Guid." achieves impressive performance. When the initial materials of prior models are not used (``w/o P-Guid."), we find that directly using the input image to guide the decoding process also improves performance. This is why the prior models use the operation of guidance feature fusion. Based on the above conclusion, the results of ``w/o Opt." and ``Full" demonstrate that the finetuning operation can further enhance the performance of material estimation. 

When calculating the rendering loss for \GeometryPriorModelShort and \AlbedoPriorModelShort, ``w/o Opt" using \ControlledPBRRenderingShort leads to higher accuracy in ``N", ``Disp", and ``D" than ``w/o Opt-C". This demonstrates that \ControlledPBRRenderingShort reduces the interference of non-optimized materials on the rendering results, thus enhancing the robustness of material estimation.

Subsurface scattering can make the reflection of the skin more realistic, and displacement can correctly represent the geometric surface, which is conducive to achieve detailed rendering. Therefore, as shown in Table~\ref{Table-comparison_on_openhumanbrdf_dataset} and Fig.~\ref{img-ablation_on_openhumanbrd_sup}, the relighting performance of ``Full" is better than that of ``w/o SSS" and ``w/o Disp", and the skin areas are more realistic.

\subsection{Applications}
\Ours can effectively estimate PBR materials from a single full-body image. These estimated materials are render-ready for Blender,  and can be applied to various applications, such as relighting and material editing.

\begin{figure}[!b]
	\centering 
	\begin{minipage}[b]{\linewidth} 
		\centering 
		\newcommand{\myvspace}{1.0 pt} 
		\newcommand{\widthOfFullPage}{1} 
		\newcommand{\widthOfMiniPage}{0.99}
		\newcommand{\format}{png}
		\subfloat{
			\begin{minipage}[b]{\widthOfFullPage\linewidth} 
				\centering
				\includegraphics[width=\widthOfMiniPage\linewidth]{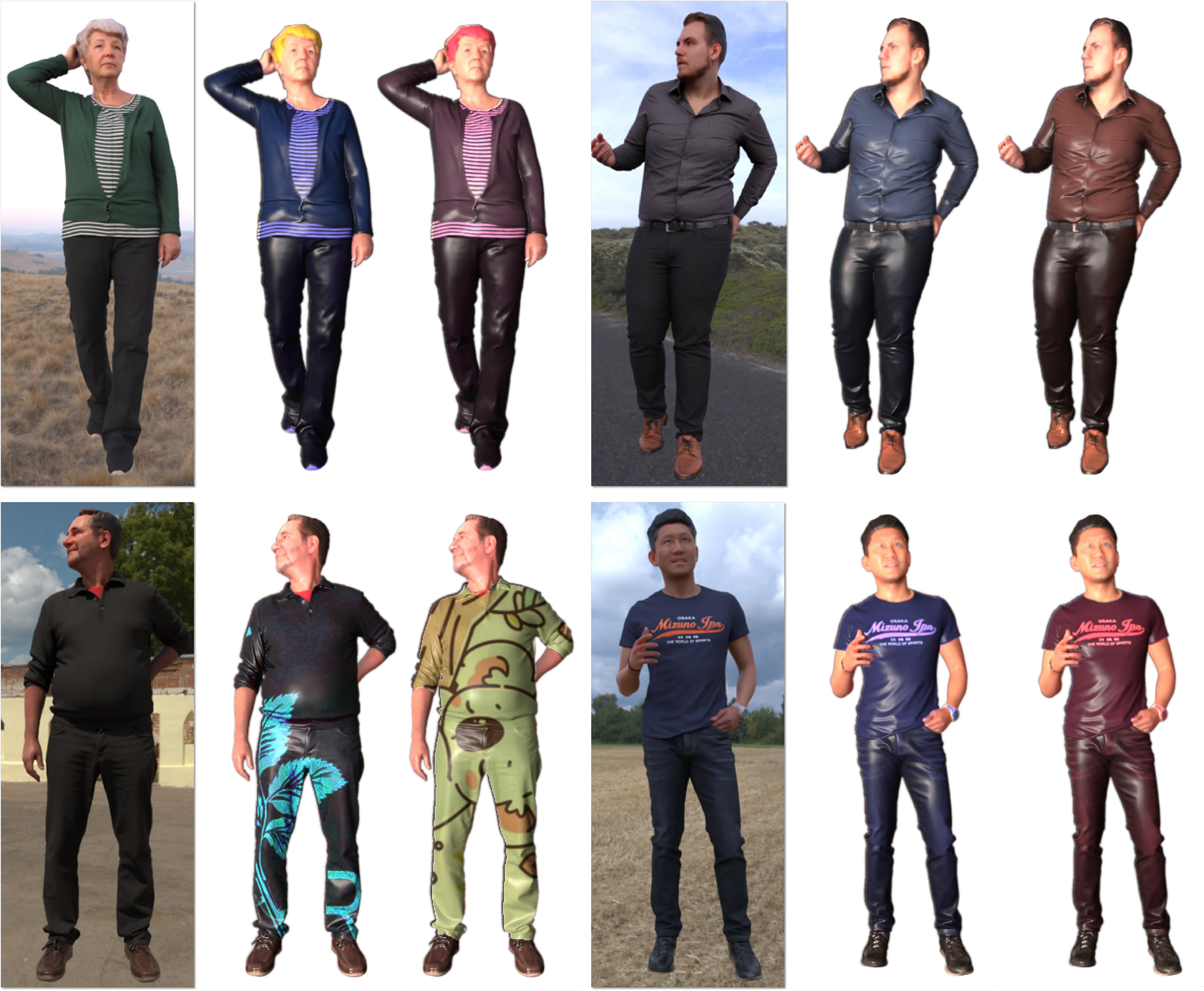}
				\begin{flushleft}
					\small
					\vspace{-0.5em}
					\hspace{1.4em}Input
					\hspace{2.1em}Edit 1
					\hspace{1.3em}Edit 2
					\hspace{2.2em}Input
					\hspace{2.4em}Edit 1
					\hspace{2.1em}Edit 2
				\end{flushleft}
			\end{minipage}
		}
	\end{minipage}
	\caption{Material Editing. Please see the dynamic editing process in the attached video.}
	\label{img-material_editing}
\end{figure}

\textbf{Material Editing.}
As shown in Fig.~\ref{img-material_editing}, we have made a class-level appearance editing. 
~Since our method can estimate PBR materials with classification, we can obtain hair, skin, fabric, and leather masks from the estimated roughness, specular albedo, or subsurface scattering. Then, we can easily edit the materials, such as changing the cloth from fabric to shiny leather, and changing the color of cloth and hair. We provide the editing demo in the attached video.
\begin{figure}[!t]
	\centering 
	\begin{minipage}[b]{\linewidth} 
		\centering 
		\newcommand{\myvspace}{1.0 pt} 
		\newcommand{\widthOfFullPage}{1} 
		\newcommand{\widthOfMiniPage}{0.99}
		\newcommand{\format}{png}
		\subfloat{
			\begin{minipage}[b]{\widthOfFullPage\linewidth} 
				\centering
				\includegraphics[width=\widthOfMiniPage\linewidth]{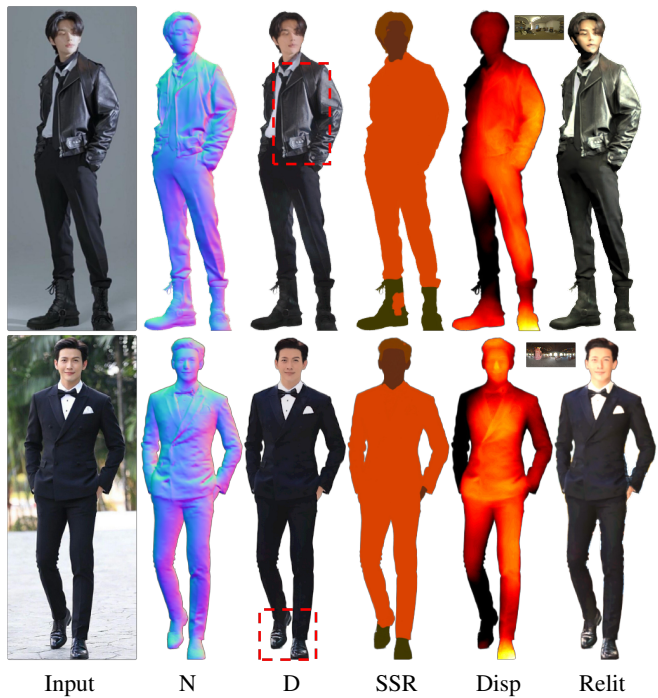}
				\begin{flushleft}
					\small
					\vspace{-0.5em}
					\hspace{1.7em}Input
					\hspace{3.2em}N
					\hspace{3.3em}D
					\hspace{2.8em}SSR
					\hspace{2.1em}Disp
					\hspace{2.1em}Relit
				\end{flushleft}
			\end{minipage}
		}
	\end{minipage}
	\caption{Limitation samples. ``N" means normal. ``D" means diffuse albedo. ``Disp" means displacement. ``Relit" means the relighting appearance under the novel environment map. ``SSR" means the result of subsurface scattering, specular albedo and roughness, which are concatenated under channel axis. }
	\label{img-limitation_highlight}
\end{figure}

\subsection{Limitations}
\label{limitation}
Our model struggles to handle samples with strong illuminations. As shown in Fig.~\ref{img-limitation_highlight}, the highlight and shadow may be baked into the diffuse albedo.
~The reason could be that the input is a single image, which contains only single view and single type of illumination. This makes it extremely difficult to decouple the material and illumination. 
A robust light-aware model or processing of the input image, such as highlight and shadow removal, is a feasible approach to fix this issue.
~The materials related to the human body in the real world are complex, our dataset considers the four most relevant materials, which limits robustness and makes it difficult to handle composite materials, such as dusty fabrics.

\section{Conclusion and Future work}
\label{conclusion}
We have presented \Ours, a method for full-body PBR material estimation from a single image.
~Our method mainly includes the following core points: (a) Construct a higher-quality Human PBR material dataset, which has realistic skin representation. (b) Propose a well-designed model (\Ours) with progressive training strategy, which employs two stages to obtain the initial materials from the prior models and finetunes it through joint optimization to obtain the final materials. (c) Based on well-designed losses, we train the model (\Ours) on our \OursDataset dataset to estimate render-ready PBR material, and employ Blender to achieve realistic relighting and material editing.
~Extensive experiments on \OursDataset dataset and real data demonstrate the superior performance of our method.

\textbf{Future Work.}
Limited by model parameters and computing resources, the resolution of PBR of our estimated materials is $512\times512$. In the future, we will explore building a more efficient network and designing an effective strategy to produce higher-resolution results.
%
%
%
%


\bibliographystyle{IEEEtran}
\normalem
\bibliography{mybib}

\begin{thebibliography}{10}
\providecommand{\url}[1]{#1}
\csname url@samestyle\endcsname
\providecommand{\newblock}{\relax}
\providecommand{\bibinfo}[2]{#2}
\providecommand{\BIBentrySTDinterwordspacing}{\spaceskip=0pt\relax}
\providecommand{\BIBentryALTinterwordstretchfactor}{4}
\providecommand{\BIBentryALTinterwordspacing}{\spaceskip=\fontdimen2\font plus
\BIBentryALTinterwordstretchfactor\fontdimen3\font minus \fontdimen4\font\relax}
\providecommand{\BIBforeignlanguage}[2]{{%
\expandafter\ifx\csname l@#1\endcsname\relax
\typeout{** WARNING: IEEEtran.bst: No hyphenation pattern has been}%
\typeout{** loaded for the language `#1'. Using the pattern for}%
\typeout{** the default language instead.}%
\else
\language=\csname l@#1\endcsname
\fi
#2}}
\providecommand{\BIBdecl}{\relax}
\BIBdecl

\bibitem{TotalRelighting_2021_Sig_single_image_human_phong_neural_render}
R.~Pandey, S.~O. Escolano, C.~Legendre, C.~Haene, S.~Bouaziz, C.~Rhemann, P.~Debevec, and S.~Fanello, ``Total relighting: Learning to relight portraits for background replacement,'' vol.~40, no.~4, August 2021.

\bibitem{Lagunas_2021_EGSR_single_image_human_neural_render}
M.~Lagunas, X.~Sun, J.~Yang, R.~Villegas, J.~Zhang, Z.~Shu, B.~Masia, and D.~Gutierrez, ``Single-image full-body human relighting,'' in \emph{Eurographics Symposium on Rendering (EGSR)}.\hskip 1em plus 0.5em minus 0.4em\relax The Eurographics Association, 2021.

\bibitem{AFHIR_Daichi_2024_arxiv}
D.~Tajima, Y.~Kanamori, and Y.~Endo, ``All-frequency full-body human image relighting,'' \emph{arXiv preprint arXiv:2411.00356}, 2024.

\bibitem{ICLight}
L.~Zhang, A.~Rao, and M.~Agrawala, ``Scaling in-the-wild training for diffusion-based illumination harmonization and editing by imposing consistent light transport,'' in \emph{The Thirteenth International Conference on Learning Representations}, 2025.

\bibitem{SwitchLight_2024_CVPR_pbr_and_neural_render}
H.~Kim, M.~Jang, W.~Yoon, J.~Lee, D.~Na, S.~Woo, and B.~Ai, ``\BIBforeignlanguage{en-US}{Switchlight: Co-design of physics-driven architecture and pre-training framework for human portrait relighting},'' \emph{\BIBforeignlanguage{en-US}{Proceedings of the IEEE/CVF Conference on Computer Vision and Pattern Recognition (CVPR)}}, pp. 25\,096--25\,106, Jan 2024.

\bibitem{RTR_2018_Akenine}
T.~Akenine-Mo and N.~Hoffman, ``Real-time rendering,'' 2018.

\bibitem{Blender}
\BIBentryALTinterwordspacing
{Blender Online Community}, \emph{Blender - a 3D modelling and rendering package}, Blender Foundation, Stichting Blender Foundation, Amsterdam, 2018. [Online]. Available: \url{http://www.blender.org}
\BIBentrySTDinterwordspacing

\bibitem{Photorealistic_2022_CVPR_single_image_human_neural_render}
T.~Alldieck, M.~Zanfir, and C.~Sminchisescu, ``Photorealistic monocular 3d reconstruction of humans wearing clothing,'' in \emph{Proceedings of the IEEE/CVF Conference on Computer Vision and Pattern Recognition (CVPR)}, 2022.

\bibitem{Unsupervised_Learning_Liu_2020_CVPR}
Y.~Liu, Y.~Li, S.~You, and F.~Lu, ``Unsupervised learning for intrinsic image decomposition from a single image,'' in \emph{Proceedings of the IEEE/CVF Conference on Computer Vision and Pattern Recognition (CVPR)}, June 2020.

\bibitem{Single_image_portrait_relighting_2019_SIGGRAPH}
T.~Sun, J.~T. Barron, Y.-T. Tsai, Z.~Xu, X.~Yu, G.~Fyffe, C.~Rhemann, J.~Busch, P.~Debevec, and R.~Ramamoorthi, ``\BIBforeignlanguage{en-US}{Single image portrait relighting},'' \emph{\BIBforeignlanguage{en-US}{ACM Transactions on Graphics (TOG)}}, p. 1–12, Aug 2019.

\bibitem{Li_2018_TOG_single_image_natural_svbrdfs}
Z.~Li, Z.~Xu, R.~Ramamoorthi, K.~Sunkavalli, and M.~Chandraker, ``Learning to reconstruct shape and spatially-varying reflectance from a single image,'' \emph{ACM Transactions on Graphics (TOG)}, vol.~37, no.~6, dec 2018.

\bibitem{NeRO_2023_Sig_Multi_images_natural_svbrdf}
Y.~Liu, P.~Wang, C.~Lin, X.~Long, J.~Wang, L.~Liu, T.~Komura, and W.~Wang, ``Nero: Neural geometry and brdf reconstruction of reflective objects from multiview images,'' \emph{ACM Transactions on Graphics (TOG)}, vol.~42, no.~4, jul 2023.

\bibitem{RADN_2018_TOG_single_image_natural_svbrdf}
V.~Deschaintre, M.~Aittala, F.~Durand, G.~Drettakis, and A.~Bousseau, ``Single-image svbrdf capture with a rendering-aware deep network,'' \emph{ACM Transactions on Graphics (TOG)}, vol.~37, no.~4, jul 2018.

\bibitem{HATSNet_2021_TOG_single_image_natural_svbrdf}
J.~Guo, S.~Lai, C.~Tao, Y.~Cai, L.~Wang, Y.~Guo, and L.-Q. Yan, ``Highlight-aware two-stream network for single-image svbrdf acquisition,'' \emph{ACM Transactions on Graphics (TOG)}, vol.~40, no.~4, jul 2021.

\bibitem{PhySG_2021_CVPR_multi_images_natural_brdf}
K.~Zhang, F.~Luan, Q.~Wang, K.~Bala, and N.~Snavely, ``{PhySG}: {I}nverse rendering with spherical gaussians for physics-based material editing and relighting,'' in \emph{Proceedings of the IEEE/CVF Conference on Computer Vision and Pattern Recognition (CVPR)}.\hskip 1em plus 0.5em minus 0.4em\relax New York, NY, USA: Association for Computing Machinery, 2021, pp. 5453--5462.

\bibitem{NeILF_2022_ECCV_Multi_images_natural_svbrdf}
Y.~Yao, J.~Zhang, J.~Liu, Y.~Qu, T.~Fang, D.~McKinnon, Y.~Tsin, and L.~Quan, ``Neilf: Neural incident light field for physically-based material estimation,'' in \emph{European Conference on Computer Vision (ECCV)}, 2022.

\bibitem{MaXiaoHe_TVCG_2024}
X.~Ma, Y.~Yu, H.~Wu, and K.~Zhou, ``Efficient reflectance capture with a deep gated mixture-of-experts,'' \emph{IEEE Transactions on Visualization and Computer Graphics (TVCG)}, vol.~30, no.~7, pp. 4246--4256, 2024.

\bibitem{BRDF_Acquisition_TVCG_2024}
E.~Miandji, T.~Tongbuasirilai, S.~Hajisharif, B.~Kavoosighafi, and J.~Unger, ``Frost-brdf: A fast and robust optimal sampling technique for brdf acquisition,'' \emph{IEEE Transactions on Visualization and Computer Graphics (TVCG)}, vol.~30, no.~7, pp. 4390--4402, 2024.

\bibitem{10929734}
J.~Li, Q.~Deng, H.~Ling, and B.~Huang, ``Dpcs: Path tracing-based differentiable projector-camera systems,'' \emph{IEEE Transactions on Visualization and Computer Graphics (TVCG)}, vol.~31, no.~5, pp. 3666--3676, 2025.

\bibitem{Relightable_2023_ICCV_video_human_svbrdf}
W.~Sun, Y.~Che, H.~Huang, and Y.~Guo, ``Neural reconstruction of relightable human model from monocular video,'' in \emph{Proceedings of the IEEE/CVF International Conference on Computer Vision}, 2023, pp. 397--407.

\bibitem{Nerfactor_2021_SigAsia_multi_images_natural_brdf}
X.~Zhang, P.~P. Srinivasan, B.~Deng, P.~Debevec, W.~T. Freeman, and J.~T. Barron, ``Nerfactor: Neural factorization of shape and reflectance under an unknown illumination,'' \emph{ACM Transactions on Graphics (TOG)}, vol.~40, no.~6, pp. 1--18, 2021.

\bibitem{Li_2017_TOG_single_image_plane_natural_svbrdf}
X.~Li, Y.~Dong, P.~Peers, and X.~Tong, ``Modeling surface appearance from a single photograph using self-augmented convolutional neural networks,'' \emph{ACM Transactions on Graphics (TOG)}, vol.~36, no.~4, jul 2017.

\bibitem{LATNet_2022_TOG_single_image_natural_svbrdf}
X.~Zhou and N.~K. Kalantari, ``Look-ahead training with learned reflectance loss for single-image svbrdf estimation,'' \emph{ACM Transactions on Graphics (TOG)}, vol.~41, no.~6, nov 2022.

\bibitem{DIR_2019_TOG_single_image_natural_svbrdf}
D.~Gao, X.~Li, Y.~Dong, P.~Peers, K.~Xu, and X.~Tong, ``Deep inverse rendering for high-resolution svbrdf estimation from an arbitrary number of images,'' \emph{ACM Transactions on Graphics (TOG)}, vol.~38, no.~4, jul 2019.

\bibitem{MaterialGAN_2020_TOG_single_image_natural_svbrdf}
Y.~Guo, C.~Smith, M.~Ha\v{s}an, K.~Sunkavalli, and S.~Zhao, ``Materialgan: Reflectance capture using a generative svbrdf model,'' \emph{ACM Transactions on Graphics (TOG)}, vol.~39, no.~6, nov 2020.

\bibitem{InvRender_2022_CVPR_Multi_images_natural_svbrdf}
Y.~Zhang, J.~Sun, X.~He, H.~Fu, R.~Jia, and X.~Zhou, ``Modeling indirect illumination for inverse rendering,'' in \emph{Proceedings of the IEEE/CVF Conference on Computer Vision and Pattern Recognition (CVPR)}.\hskip 1em plus 0.5em minus 0.4em\relax New York, NY, USA: Association for Computing Machinery, 2022, pp. 18\,643--18\,652.

\bibitem{Relightify_2023_ICCV_DiffusinModel_SVBRDF_facial}
F.~Paraperas~Papantoniou, A.~Lattas, S.~Moschoglou, and S.~Zafeiriou, ``Relightify: Relightable 3d faces from a single image via diffusion models,'' in \emph{Proceedings of the IEEE/CVF International Conference on Computer Vision (ICCV)}, 2023.

\bibitem{TensoIR_2023_CVPR_Multi_images_natural_svbrdf}
H.~Jin, I.~Liu, P.~Xu, X.~Zhang, S.~Han, S.~Bi, X.~Zhou, Z.~Xu, and H.~Su, ``Tensoir: Tensorial inverse rendering,'' in \emph{Proceedings of the IEEE/CVF Conference on Computer Vision and Pattern Recognition (CVPR)}, 2023.

\bibitem{TenseRF_2022_ECCV}
A.~Chen, Z.~Xu, A.~Geiger, J.~Yu, and H.~Su, ``Tensorf: Tensorial radiance fields,'' in \emph{European Conference on Computer Vision (ECCV)}, 2022.

\bibitem{NeRD_2021_ICCV_Multi_images_natural_svbrdf}
M.~Boss, R.~Braun, V.~Jampani, J.~T. Barron, C.~Liu, and H.~P. Lensch, ``Nerd: Neural reflectance decomposition from image collections,'' in \emph{IEEE International Conference on Computer Vision (ICCV)}, 2021.

\bibitem{Nvdiffrec_2022_CVPR_natural_multi_images_svbrdf}
J.~Munkberg, J.~Hasselgren, T.~Shen, J.~Gao, W.~Chen, A.~Evans, T.~M\"uller, and S.~Fidler, ``{Extracting Triangular 3D Models, Materials, and Lighting From Images},'' in \emph{Proceedings of the IEEE/CVF Conference on Computer Vision and Pattern Recognition (CVPR)}, June 2022, pp. 8280--8290.

\bibitem{Factored_NeuS_2023_arxiv_Multi_images_natural_svbrdf}
Y.~Fan, I.~Skorokhodov, O.~Voynov, S.~Ignatyev, E.~Burnaev, P.~Wonka, and Y.~Wang, ``Factored-neus: Reconstructing surfaces, illumination, and materials of possibly glossy objects,'' \emph{arXiv preprint arXiv:2305.17929}, 2023.

\bibitem{AvatarMe++_2022_TPAMI_Lattas}
A.~Lattas, S.~Moschoglou, S.~Ploumpis, B.~Gecer, A.~Ghosh, and S.~Zafeiriou, ``Avatarme++: Facial shape and brdf inference with photorealistic rendering-aware gans,'' \emph{IEEE Transactions on Pattern Analysis and Machine Intelligence}, vol.~44, no.~12, pp. 9269--9284, 2021.

\bibitem{AvatarMe_2020_CVPR_Lattas}
A.~Lattas, S.~Moschoglou, B.~Gecer, S.~Ploumpis, V.~Triantafyllou, A.~Ghosh, and S.~Zafeiriou, ``Avatarme: Realistically renderable 3d facial reconstruction" in-the-wild",'' in \emph{Proceedings of the IEEE/CVF Conference on Computer Vision and Pattern Recognition (CVPR)}, 2020, pp. 760--769.

\bibitem{Shu_Hadap_Shechtman_Sunkavalli_Paris_Samaras_2017}
Z.~Shu, S.~Hadap, E.~Shechtman, K.~Sunkavalli, S.~Paris, and D.~Samaras, ``\BIBforeignlanguage{en-US}{Portrait lighting transfer using a mass transport approach},'' \emph{\BIBforeignlanguage{en-US}{ACM Transactions on Graphics (TOG)}}, p.~1, Jul 2017.

\bibitem{Relit_NeuLF_2023_ACMMM_Multi_images_facial_svbrdf}
Z.~Li, L.~Song, C.~Liu, J.~Yuan, and Y.~Xu, ``Relit-neulf: Efficient novel view synthesis with neural 4d light field,'' in \emph{Proceedings of the 31st ACM International Conference on Multimedia}, 2023.

\bibitem{HighRes_2023_CVPR_images_facial_svbrdf}
D.~Azinovi\'c, O.~Maury, C.~Hery, M.~Nie{\ss}ner, and J.~Thies, ``High-res facial appearance capture from polarized smartphone images,'' in \emph{Proceedings of the IEEE/CVF Conference on Computer Vision and Pattern Recognition (CVPR)}, June 2023.

\bibitem{Relighting4D_2022_ECCV_video_human_svbrdf}
Z.~Chen and Z.~Liu, ``Relighting4d: Neural relightable human from videos,'' in \emph{European Conference on Computer Vision}.\hskip 1em plus 0.5em minus 0.4em\relax Springer, 2022, pp. 606--623.

\bibitem{zhen_2023_Arxiv_video_human_svbrdf}
Z.~Xu, S.~Peng, C.~Geng, L.~Mou, Z.~Yan, J.~Sun, H.~Bao, and X.~Zhou, ``Relightable and animatable neural avatar from sparse-view video,'' in \emph{Proceedings of the IEEE/CVF Conference on Computer Vision and Pattern Recognition (CVPR)}, 2024, pp. 990--1000.

\bibitem{HDRI}
PolyHaven, ``{HDR} environment images,'' \url{https://polyhaven.com/hdris}.

\bibitem{RenderPeople}
{RenderPeople Team}, ``Renderpeople,'' https://renderpeople.com, 2021.

\bibitem{Rembg}
\BIBentryALTinterwordspacing
D.~Gatis, 2020. [Online]. Available: \url{https://github.com/danielgatis/rembg}
\BIBentrySTDinterwordspacing

\bibitem{DisneyBRDF_2012_burley}
B.~Burley and W.~D.~A. Studios, ``Physically-based shading at disney,'' in \emph{Acm Siggraph}, vol. 2012.\hskip 1em plus 0.5em minus 0.4em\relax vol. 2012, 2012, pp. 1--7.

\bibitem{Tensorflow_2016_Abadi}
M.~Abadi, P.~Barham, J.~Chen, Z.~Chen, A.~Davis, J.~Dean, M.~Devin, S.~Ghemawat, G.~Irving, M.~Isard, M.~Kudlur, J.~Levenberg, R.~Monga, S.~Moore, D.~G. Murray, B.~Steiner, P.~Tucker, V.~Vasudevan, P.~Warden, M.~Wicke, Y.~Yu, and X.~Zheng, ``Tensorflow: a system for large-scale machine learning,'' ser. OSDI'16.\hskip 1em plus 0.5em minus 0.4em\relax USA: USENIX Association, 2016, p. 265–283.

\end{thebibliography}

\end{document}